\documentclass[11pt,letterpaper]{article}
\usepackage{fullpage}

\usepackage{amsmath}
\usepackage{amssymb}
\usepackage{graphicx}
\usepackage[semicolon]{natbib}

\usepackage[labelformat=simple]{subcaption}
\usepackage[noend]{algorithmic}

\usepackage{color}
\usepackage[normalem]{ulem}

\newcommand{\todo}[1]{}
\newcommand{\review}[1]{#1}

\usepackage{macros}
\newif\iftruegibbs
\truegibbsfalse
\newif\iffalseneg
\falsenegfalse
\newif\iffactored
\factoredfalse

\begin{document}

\title{Object-based World Modeling in Semi-Static Environments \\ with Dependent Dirichlet-Process Mixtures}
\author{Lawson L.S. Wong, Thanard Kurutach, Leslie Pack Kaelbling, Tom\'as Lozano-P\'erez \\
CSAIL, MIT, Cambridge, MA 02139 \\
\{ \tt{lsw, kurutach, lpk, tlp} \}@csail.mit.edu}
\date{}
\maketitle

\begin{abstract}
To accomplish tasks in human-centric indoor environments,
robots need to represent and understand the world in terms of objects and their attributes.
We refer to this attribute-based representation as a world model,
and consider how to acquire it via noisy perception and maintain it over time, as objects are added, changed, and removed in the world.
Previous work has framed this as multiple-target tracking problem, where objects are potentially in motion at all times.
Although this approach is general, it is computationally expensive.
We argue that such generality is not needed in typical world modeling tasks, where objects only change state occasionally.
More efficient approaches are enabled by restricting ourselves to such semi-static environments.

We consider a previously-proposed clustering-based world modeling approach that assumed static environments,
and extend it to semi-static domains by applying a dependent Dirichlet-process (DDP) mixture model.
We derive a novel MAP inference algorithm under this model, subject to data association constraints.
We demonstrate our approach improves computational performance in semi-static environments.
\end{abstract}

\section{Introduction}
\label{sec.introduction}
\todo{Make this flow from previous chapter.}

There are many situations in which it is important for an automated
system to maintain an estimate of the state of a complex dynamical
system.  Many physical systems are well described in terms of a set of
objects, attributes of those objects, and relations between them.  The
number and properties of the objects in the world may change over
time, and they are only partially observable due to noise and
occlusion in the observation process.  Domains that are appropriately
modeled this way include:  a household robot, which must maintain an
estimate of the contents of a refrigerator that is used by multiple
other people based on partial views of its contents;  a
wildlife-monitoring drone, which must maintain an estimate of the
number, age, and health of elephants in a herd based on a sequence of
photos of the herd moving through a forest;  a surveillance satellite,
which must estimate the number, activity, and hostility of soldiers in
an enemy camp based on photos capturing people only when they are
outside of buildings.
\todo{Semi-static state examples?}

Estimating properties of individuals from noisy observations is a
relatively simple statistical estimation problem if the observations
are labeled according to which individual generated them.  Even when
the underlying attributes of the individual change over time, the
problem of inferring the history of each individual's attributes can be
reduced to a problem of inference in a hidden Markov model.    

The key difficulty in the problems described above is \emph{data association}.
We do not know which particular individual is
responsible for each observation and so determining an appropriate
association of observations to individuals is key.
The only information we have to help make such associations are noisy and partial observations,
which may contain errors both in attribute values and in number.

Within the context of world modeling, \citet{Cox1994} first identified this issue,
and applied well-known multiple-hypothesis tracking (MHT) methods to resolve
the issue \citep{Reid1979,BarShalom1988,Elfring2013}.
\todo{Refer back to static case for what an MHT does.}
Recently, \citet{Oh2009} have pointed out drawbacks in using the MHT,
which include inefficiency due to considering an exponential number of hypotheses,
and the inability to revisit associations from previously-considered views
(the MHT is essentially a forward filtering algorithm).
Inspired by this, they and others \citep{Dellaert2003,Pasula1999}
have proposed different Markov-chain Monte Carlo (MCMC) methods for data association.
See \todo{Section~\ref{todo} in Chapter~\ref{todo}} \citet{Wong2015} for in-depth coverage about
previous work in semantic world modeling and data association.

The methods mentioned above were all formulated for multiple-target tracking problems,
where the each target's state (typically location) changes between observations.
However, if we consider applications such as tracking objects in a household,
the dynamics are typically different: most objects tend to stay in the same state
when they are not being actively used.
In this paper, we study the world modeling problem in \emph{semi-static environments},
where time is divided into known epochs, and within each epoch the world is stationary.
It seems intuitive that data association should be easier within static periods,
since there is no uncertainty arising from stochastic dynamics.


\begin{figure*}[t!]
\begin{center}
\includegraphics[width=\linewidth]{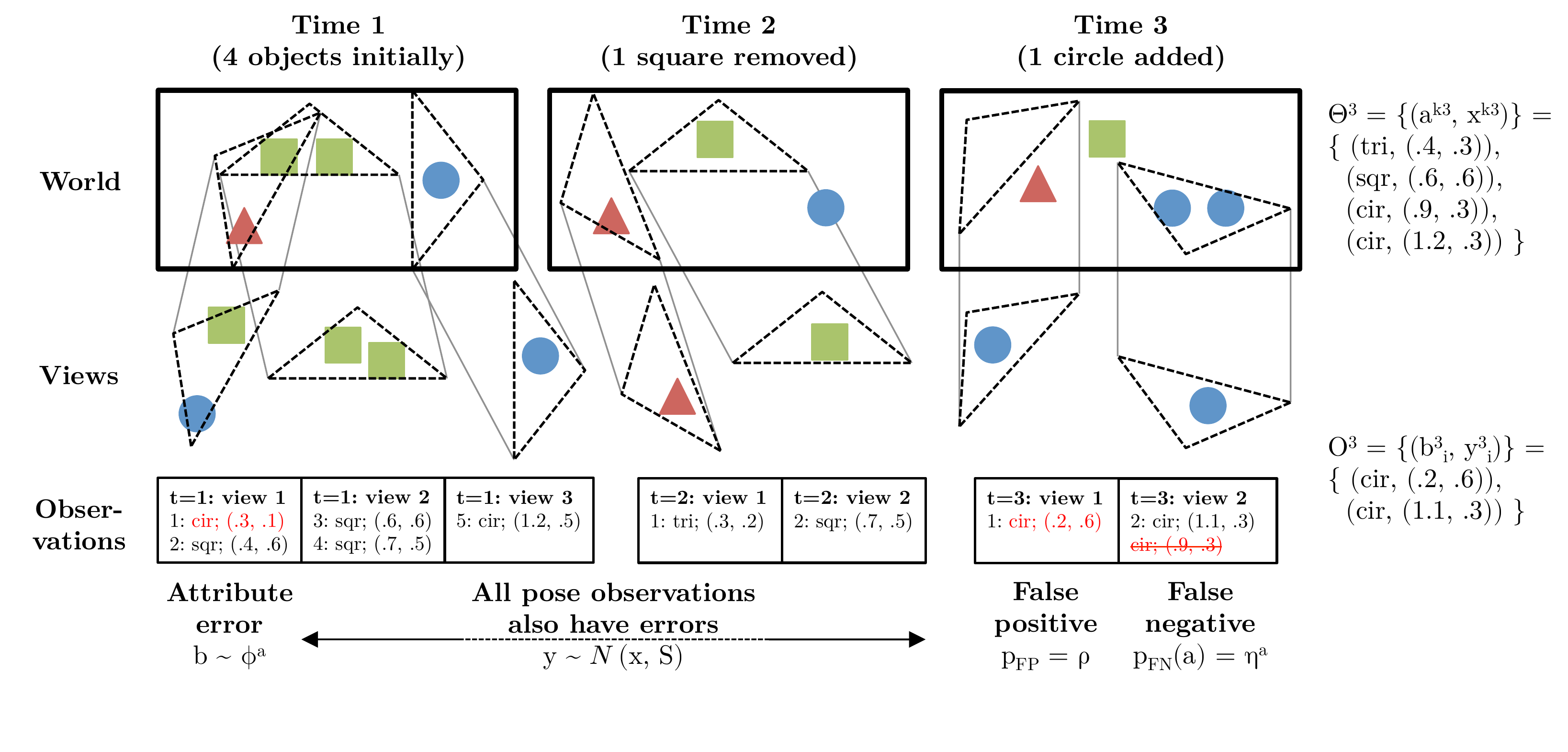}
\end{center}
\caption[An illustration of the semantic world modeling problem in a semi-static world.]
{An illustration of the world modeling problem.
An unknown number of objects exist in the world (top row), and change in pose and number over time
(world at each epoch enclosed in box). At each epoch, limited views of the world are captured,
as depicted by the triangular viewcones. Within these viewcones,
objects and their attributes are detected using black-box perception modules
(e.g., off-the-shelf object detectors).
In this example, the attributes are shape type (discrete) and 2-D location.
The observations are noisy, as depicted by the perturbed versions of viewcones in the middle row.
Uncertainty exists both in the attribute values and the existence of objects,
as detections may include false positives and negatives (e.g., $t=3$).
The actual attribute detection values obtained from the views are shown
in the bottom row (``Observations''); this is the format of input data.
Given these noisy measurements as input, the goal is to determine which objects were in existence
at each epoch, their attribute values (e.g., $\Param^3$ in top right),
and their progression over time.}
\label{fig.worldmodel}
\end{figure*}

An alternative approach to data association is to perform inference
over the entire time-series of observations and to think of it as a
problem of clustering:  we wish to group together similar detections
over time, under the assumption that they will have been generated by
the same individual.  Bayesian nonparametric models, such as the
Dirichlet-process mixture model (DPMM), can be used to model domains
in which the number of individuals is unknown \textit{a priori};
in \todo{the previous chapter} \citet{Wong2015}, we found that a
state-estimation technique based on DPMM clustering was
effective for determining the number and type of objects in a static
domain, given a sequence of images with partial views of the
scene and significant occlusion.

In this \todo{chapter} paper, we apply the clustering approach to the much more
difficult case of a dynamic domain in which the attributes of objects
may change over time, new objects may appear, and old objects may
permanently disappear.  The DPMM is not an appropriate model for this
problem, but an extension, the \emph{dependent Dirichlet process
mixture model} (DDPMM), which models dependencies between a collection
of clusters, can be used effectively.
In particular, we use a construction proposed by \citet{Lin2010}
for a class of DDPs that can be represented as a Markov chain over DPs.
In our case of semi-static world modeling,
we model objects in each static epoch as clusters in a DPMM,
and clusters between epochs are related by Markovian transitions,
thus forming a DDPMM.

In the remainder, we will formalize the world modeling problem,
review the DDP construction and apply it to our problem,
and derive a novel maximum \textit{a posteriori} (MAP) inference
algorithm for the model.
We show that this model yields computational advantages
for tracking in semi-static environments,
both in simulation and on real-world data.


\section{Problem Definition}

In world modeling, we seek the state of the world,
consisting an unknown finite number $\numcluster^\timecurr$ of objects, which changes over time.
Object $\cluster$ at epoch $\timecurr$ has attribute values $\beldef{\param}$.
We sometimes decompose $\beldef{\param}$ into $\parens{\beldeffixed{\attr}, \beldef{\pose}}$,
where $\attr$ is a vector of fixed attributes,
and $\pose$ is a vector of attributes that may change between epochs.
The top row in Figure~\ref{fig.worldmodel} illustrates the world state over three epochs for a simple domain.


Our system obtains noisy, partial views of the world.
Each view $\view$ produces a set of observations $\worlddefview{\Obs} = \braces{\worlddefviewind{\obs}}$,
where $\worlddefviewind{\obs} = \parens{\worlddefviewind{\attrobs}, \worlddefviewind{\poseobs}}$,
corresponding to the fixed attributes $\attr$ and dynamic attributes $\pose^\timecurr$
of some (possibly non-existent) object\footnote{
Superscripts in variables will generally refer to the `context',
such as object index $\cluster$ and time index $\timecurr$.
Subscripts refer to the index in a list, such as $\worlddef{\obs} =$
$i$'th observation at time $\timecurr$.}.
Each view is also associated with a field of view $\reg$.
The collection of views in a single epoch may fail to cover the entire world.
The partial views and noisy observations are illustrated
in the middle and bottom rows of Figure~\ref{fig.worldmodel}.



The world modeling problem can now be defined:
Given observations $\Obs = \braces{\worlddefviewind{\obs}}_{(\timecurr, \view, \defind)}$
and fields of view $\braces{\reg}_{(\timecurr, \view)}$,
determine the state of objects over time $\Param = \braces{\beldef{\param}}_{(\cluster, \timecurr)}$.
The state includes not only objects' attribute values,
but also the total number of objects that existed at each epoch,
and implicitly when objects were added and removed (if at all).

There is no definitive information in the observations that will allow us to know
which particular observations correspond with which underlying objects in the world,
or even how many objects were in existence at any time step.
For example, in the views of $\timecurr=1$ shown in Figure~\ref{fig.worldmodel},
the square detected in the left-most view may correspond to either (or neither)
square in the center view. 
Also, despite there being only four objects in the world,
there were five observations because of overlapping visible regions.


The critical piece of information that is missing is the \emph{association} $\worlddefviewind{\assn}$
of an observation $\worlddefviewind{\obs}$ to an underlying object $k$.
With this information, we can perform statistical aggregation of the observations
assigned to the same object to recover its state.
We will model the associations $\Assn = \braces{\worlddefviewind{\assn}}_{(\timecurr, \view, \defind)}$
as latent variables in a Bayesian inference process.

\subsection{Observation noise model}

The observation model describes how likely an observation
$\obs = \parens{\attrobs, \poseobs}$ was generated from
some given object state $\param = \parens{\attr, \pose}$ (if any),
given by the probability $\pdfdensity{\obs}{\param}$.
For a single object, let $\param_c$ and $\param_d$ be the
true continuous and discrete attribute values respectively,
and likewise $\obs_c$ and $\obs_d$ for a single observation of the object.
We typically consider observation noise models of the following form:
\begin{align}
\label{eqn.obsnoise}
\pdfdensity{\obs}{\param} = \attrnoisefn{\param_d}{\obs_d} \;
\pdfnormal{\obs_c}{\param_c, \sensecov}
\end{align}
Here $\attrnoise$ represents a discrete confusion matrix,
where $\attrnoisefn{\param_d}{\obs_d}$ is the probability of observing $\obs_d$
given the true object has discrete attributes $\param_d$.
The continuous-valued observation $\obs_c$ is the true value $\param_c$
corrupted with zero-mean Gaussian noise, with fixed sensing covariance $\sensecov$.
The noise on $\obs_c$ and $\obs_d$ are assumed to be independent for simplicity.

Besides errors in attribute values, Figure~\ref{fig.worldmodel} also illustrates
cases of false positives and false negatives. A false positive occurs when
the observation did not originate from any true object.
We assume that this occurs at a fixed rate $\pfp$, depending on the perception system.
\review{When this occurs, $\obs_d$ has noise distribution $\attrnoise^0$,
and $\obs_c$ is uniformly distributed over the field of view $\region$.}
A false negative occurs when an object is within the sensor's field of view
but failed to be detected. We assume that an object within the field of view $\region$
will be undetected with an attribute-dependent probability $\pfn(\param)$.

\subsection{Additional assumption: Cannot-link constraint (CLC)}
\label{sec.clc}

Finally, there is an additional common domain assumption
in target-tracking problems that is essential:
within a single view, each visible object can generate at most one detection
\cite{BarShalom1988}. This implies that within a view,
each observation must be assigned to a different hypothesized underlying object.
Adopting the terminology of clustering, we refer to this as a `'cannot-link constraint'' (CLC).
The constraint is powerful because it can reduce ambiguities when there are
similar nearby objects. However, clustering algorithms typically cannot handle
such constraints, and similar to the DPMM-based data association work of
\citet{Wong2015}, we will need to modify the DDP model and inference algorithms
to handle the world modeling problem.

\section{A Clustering-Based Approach}
\label{sec.clustering}

We now specify a prior on how likely an assignment to a cluster is,
and how clusters change over time.
Since the number of clusters are unknown,
we chose to use Bayesian nonparametric mixture models,
which allow for an indefinite and unbounded number of mixture components.
(although the number of instantiated components is limited by the data size).

The Dirichlet process (DP) (\citet{Teh2010} provides a good overview),
and its application to mixture modeling \citep{Antoniak1974,Neal2000},
is a widely-studied prior for density estimation and clustering.
The DP's popularity stems from its simplicity and elegance.
However, one major limitation is that clusters cannot change over time,
a consequence of the fact that observations are assumed to be fully exchangeable.
This assumption is violated for problems like ours,
where the observed entities change over time and space.
Indeed, the previous application of DPs to world modeling mentioned above
required that the world is static, which is a significant limitation.
Various generalizations of the DP that model temporal dynamics
have thus been proposed \citep{Zhu2005,Ahmed2008}.
\todo{Write more about the other related approaches.
Include \citet{Neiswanger2014,Luo2015}.}

Many of these generalizations belong to a broad class of stochastic models
known as dependent Dirichlet processes (DDP) \citep{MacEachern1999,MacEachern2000}.
We will adopt a theoretically-appealing instance of the DDP,
based on a recently-proposed Poisson-process construction \citep{Lin2010,Lin2012}.
This construction subsumes a number of existing algorithmically-motivated DP generalizations.
Additionally, Lin's construction has the nice property that
at each time slice, the prior over clusters is marginally a DP.
Given a DP prior at time $\timecurr$, the construction specifies a dependent prior
at time $\timecurr+1$ (or another future time), which is shown to also be a DP.
The construction therefore generates a Markov chain of DPs over time,
\review{which reflects temporal dynamics between epochs in our problem}.

We now state one result of the DDP construction; see
Appendix~\ref{sec.appendix-ddp}, and \citet{Lin2012} for details.
The construction results in the following prior on parameter $\param^\timecurr$
(to be assigned to a new observation),
given past parameters $\Param^{< \timecurr}$ and parameters $\Param^\timecurr$
corresponding to clusters that have already been instantiated at the current epoch:
\begin{align}
\label{eqn.ddp}
\param^0 \;|\; \Param^0 &\propto \; \dpconc \pdfbase{\param^0} +
\sum_\cluster \numcount^{\cluster 0} \; \II \brackets{\param^0 = \param^{\cluster 0}} \\
\param^\timecurr \;|\; \Param^{\leq \timecurr} &\propto \;
\dpconc \pdfbase{\param^\timecurr} +
\sum_{\cluster: \beldef{\numcount} > 0} \numcount^{\cluster, \leq \timecurr} \;
\II \brackets{\param^\timecurr = \beldef{\param}}
+ \sum_{\cluster: \beldef{\numcount} = 0} \survive(\param^{\cluster, \timecurr-1}) \;
\numcount^{\cluster, <  \timecurr} \;
\pdftrans{\param^\timecurr}{\param^{\cluster, \timecurr-1}} \nonumber
\end{align}
At the initial time step, clusters are formed as in a standard DPMM
with concentration parameter $\dpconc$ and base distribution $\dpbase$.
For later time steps, the prior distribution on $\param$ is defined recursively.
The first two terms are similar to the base case,
for new clusters and already-instantiated clusters (in the current epoch) respectively.
The third term corresponds to previously-existing clusters
that may be removed with probability $(1 - q(\param^{\cluster, \timecurr-1}))$,
and, if it survives, is moved with transition probability
$\pdftrans{\cdot}{\param^{\cluster, \timecurr-1}}$.
$\numcount^{\cluster,\leq t}$ is the number of points that have been assigned
to cluster $\cluster$, for all time steps up to time $\timecurr$.
\review{This term is similar to that in the DP.}
Note that if $\survive \equiv 1$ and
$\pdftrans{\cdot}{\param} = \delta_{\param}$,
then the model is static, and Equation~\ref{eqn.ddp}
is equivalent to the predictive distribution in the DP.

\subsection{Inference by forward sampling}

As mentioned in the problem definition,
our focus will be on determining latent assignments
$\Assn = \braces{\worlddef{\assn}}$ of observations $\Obs = \braces{\worlddef{\obs}}$
to clusters with parameters $\Param = \braces{\beldef{\param}}$.
In the generic DDP, views do not exist yet;
those will be introduced in Section~\ref{sec.constraints}.
One way to explore the distribution of assignments is to sample repeatedly
from the assignment's conditional distribution, given all other assignments
$\excludedef{\Assn} \triangleq \Assn \setminus \braces{\worlddef{\assn}}$:
\begin{align}
\label{eqn.inference}
\PP \parensmid{\worlddef{\assn} = \cluster}{\worlddef{\obs}, \Param, \excludedef{\Assn}}
&= \integral{\probcond{\worlddef{\assn} = \cluster, \param}{\worlddef{\obs}, \Param, \excludedef{\Assn}}}{\param}{} \nonumber \\
&\propto \integral{\probcond{\worlddef{\obs}}{\param} \probcond{\param = \beldef{\param}}{\Param, \excludedef{\Assn}}}{\param}{}
\end{align}
The first term in the integrand is given by the observation noise model 
(Equation~\ref{eqn.obsnoise}), and the second term is 
given by the DDP prior (Equation~\ref{eqn.ddp}).
If $\beldef{\param}$ already exists, then $\probcond{\param}{\Param, \excludedef{\Assn}} = \II \brackets{\param = \beldef{\param}}$,
and the integrand only has support for $\param = \beldef{\param}$.
Otherwise, we have to consider all possible settings of $\beldef{\param}$,
which has a prior distribution given by Equation~\ref{eqn.ddp}.
The expression in Equation~\ref{eqn.inference} above can be decomposed into three cases,
corresponding to terms in Equation~\ref{eqn.ddp}:
\begin{align}
\label{eqn.forward}
\PP \parensmid{\worlddef{\assn} = \cluster}{\worlddef{\obs}, \Param^{\leq \timecurr}, \excludedef{\Assn^{\leq \timecurr}}} \propto
&\begin{cases}
\excludedef{\numcount^{\cluster, \leq \timecurr}}
& \pdfdensity{\worlddef{\obs}}{\beldef{\param}} \;,
\\ & \cluster \text{ existing, instantiated at } \timecurr \\[1.5ex]
\survivegen(\beldefprev{\param}) \; \excludedef{\numcount^{\cluster, < \timecurr}}
& \integral{\pdfdensity{\worlddef{\obs}}{\param} \; \pdftransgen{\param}{\beldefprev{\param}}}{\param}{} \;,
\\ & \cluster \text{ existing, not instantiated at } \timecurr \\[1.5ex]
\dpconc
& \integral{\pdfdensity{\worlddef{\obs}}{\param} \; \pdfbase{\param}}{\param}{} \;,
\\ & \cluster \text{ new}
\end{cases}
\end{align}

In the DDPMM, clusters move around the parameter space during their lifetimes,
and, depending on our chosen viewpoints, may not generate observations at some epochs.
When cluster $\cluster$ has at least one time-$\timecurr$ observation assigned to it,
it becomes \textit{instantiated} at time $\timecurr$.
Any subsequent observations at time $\timecurr$ that are assigned to cluster $\cluster$
must then share the same parameter $\beldef{\param}$;
this corresponds to the first case.
The second case is for clusters not yet instantiated at time $\timecurr$,
and we must infer $\beldef{\param}$
from the last known parameter for cluster $\cluster$, at time $\timeprev < \timecurr$.
If $\timecurr - \timeprev > 1$,
we use generalized survival and transition expressions for our application:
\begin{align}
\label{eqn.transgen}
\survivegen(\beldefprev{\param}) &\triangleq \brackets{\survive(\beldefprev{\param})}^{\timecurr - \timeprev} \\
\pdftransgen{\beldef{\param}}{\beldefprev{\param}} &= \II \brackets{\beldef{\attr} = \beldefprev{\attr}} \; \pdfnormal{\beldef{\pose}}{\beldefprev{\pose}, (\timecurr - \timeprev) \transcov(\attr^\cluster)} \nonumber
\end{align}
The third case is for new clusters that are added at time $\timecurr$.
The first and third cases essentially have the same form as the Gibbs sampler for the (static) DP
\todo{(see Equation~\ref{eqn.todo})}.

In general, since the cluster parameters $\Param$ are also unknown,
inference schemes need to alternate between sampling the cluster assignments (given parameters) as above,
and sampling the parameters given the cluster assignments.
The conditional distribution of each cluster's parameters $\braces{\beldef{\param}}$
(for each cluster $\cluster$, a sequence of parameters) can be found using Bayes' rule:
\begin{align}
\probcond{\braces{\beldef{\param}}}{\Obs, \Assn} &=
\probcond{\braces{\beldef{\param}}}{\left. \Obs \right|_{\assn = \cluster}, \Assn}
\propto \brackets{\prod_{\worlddef{\assn} = \cluster} \probcond{\worlddef{\obs}}{\beldef{\param}}} \;
\prob \parens{\braces{\beldef{\param}}}
\end{align}
Depending on the choice of parameter priors and observation functions,
the resulting conditional distributions can potentially be complicated to represent
and difficult to sample from.
With additional assumptions that will be presented next,
we can find the parameter posterior distribution efficiently
and avoid sampling the parameter entirely by ``collapsing'' it.

\subsection{Application of DDPs to world modeling}
\label{sec.ddp-worldmodel}

We now apply the DDP mixture model (DDPMM) to our semi-static world modeling problem.
For concreteness and simplicity, we consider an instance of the
world modeling problem where the fixed attribute $\attr$ is the discrete object type
(from a finite list of known types),
and the dynamic attribute $\pose$ is the continuous pose in $\mathbb{R}^d$
(either 3-D location or 6-D pose).
Despite these restrictions, our model and derivations below can be
immediately applied to problems with any fixed attributes,
and with any dynamic continuous attributes with linear-Gaussian dynamics.
\review{Arbitrary dynamic attributes can be represented in our model,
but inference will likely be more challenging because
in general we will not obtain closed-form expressions.}

For our instance of the DDPMM, we assume:
\begin{itemize}
\item Time steps in the DDP correspond to epochs in world modeling.
This implies that each epoch is modeled as a static DPMM,
\review{similar to the problem in \citet{Wong2015} \todo{the previous chapter}}.
\item The survival rate only depends on the fixed attribute,
i.e., $\survive(\param) = \survive(\attr)$.
(For us, that means the likelihood of object removal is
dependent on the object type but not its pose.)
\item Likewise, the detection probability only depends on the fixed attribute,
i.e., $\pfn(\param) = \pfn(\attr)$.
\item The dynamic attribute (pose) follows a random walk with
zero-mean Gaussian noise that depends on $\attr$
(e.g., a mug likely travels farther per epoch than a table):
\begin{align}
\pose^{\timecurr+1} = \pose^\timecurr + w, \text{ where } w \sim \normal \parens{0, \transcov(\attr)}
\end{align}
This implies that the full transition distribution (of both object type and pose) is:
\begin{align}
\label{eqn.worldmodeltrans}
\pdftrans{\param^{\timecurr+1}}{\param^\timecurr} = \II \brackets{\attr^{\timecurr+1} = \attr^\timecurr} \; \pdfnormal{\pose^{\timecurr+1}}{\pose^\timecurr, \transcov(\attr)}
\end{align}
\item At each epoch, the DP base distribution has the following form:
\begin{align}
\label{eqn.worldmodelbase}
\pdfbase{\param} \triangleq \attrprior(\attr) \; \pdfnormal{\pose}{\belmean^0, \belcov^0}
\end{align}
Here a (discrete) prior $\attrprior$ over the object type,
and a normal distribution over the object pose.
The initial covariance $\belcov^0$ is large, in order to give
reasonable likelihood of an object being introduced at any location.
In fact, we will set $\belcov^0 = \infty I$ and $\belmean^0 = \mathbf{0}$,
representing a noninformative prior over the location.
Details can be found in Appendix~\ref{sec.appendix-worldmodel}.
\end{itemize}

The above choices for the dynamics and base distribution implies that
the parameter posterior and predictive distributions have closed-form expressions.
The posterior distribution of the dynamic attribute is a mixture of Gaussians,
with a component for each possible value of the fixed attribute $\attr$
(since the process noise $\transcov(\attr)$ may be different),
weighted by the posterior probability of $\attr$.
\todo{Verify this.}
In practice, we track the pose using only the dynamics of the most-likely object type.
Thus, in our application, each cluster will maintain a discrete posterior distribution
$\belpmf(\attr)$ for the object type,
and a single Kalman filter / Rauch-Tung-Striebel (RTS) smoother \todo{(cite?)}
for the object pose distribution.
The latter is represented as a sequence of means and covariances
$\braces{\belmean^\timecurr, \belcov^\timecurr}_{\timecurr = \timebirth}^{\timedeath}$
over the cluster's lifetime $\timecurr \in \brackets{\timebirth, \timedeath}$,
with the interpretation that
$\pose^\timecurr \sim \normal \parens{\belmean^\timecurr, \belcov^\timecurr}$.

As mentioned previously, because we have compact representations of the
parameter posterior distributions, we can analytically integrate them out sampling them.
We first modify the forward sampling equation (Equation~\ref{eqn.forward})
to reflect this ``collapsing'' operation.
Since we can no longer condition on the parameters themselves,
we instead need to condition on the other observations $\excludedef{\Obs}$
and their current cluster assignments $\excludedef{\Assn}$,
and use posterior \emph{predictive} likelihoods of the form
$\probcond{\worlddef{\obs}}{\excludedef{\Obs}^\cluster}$
to evaluate the current observation $\worlddef{\obs}$:
\begin{align}
\label{eqn.forward-collapsed}
\PP &\parensmid{\worlddef{\assn} = \cluster}{\worlddef{\obs}, \excludedef{\Obs^{\leq \timecurr}}, \excludedef{\Assn^{\leq \timecurr}}}
\propto \probcond{\worlddef{\obs}}{\worlddef{\assn} = \cluster, \excludedef{\Obs^{\leq \timecurr}}, \excludedef{\Assn^{\leq \timecurr}}} \;
\probcond{\worlddef{\assn} = \cluster}{\excludedef{\Obs^{\leq \timecurr}}, \excludedef{\Assn^{\leq \timecurr}}} \nonumber \\
&\propto \integral{\brackets{\probcond{\worlddef{\obs}}{\beldef{\param}} \;
\probcond{\beldef{\param}}{\excludedef{\Obs^{\cluster, \leq \timecurr}}}} \;
\probcond{\worlddef{\assn} = \cluster}{\excludedef{\Assn^{\leq \timecurr}}}}{\beldef{\param}}{} \nonumber \\
&\propto
\begin{cases}
\excludedef{\numcount^{\cluster, \leq \timecurr}}
& \integral{\probcond{\worlddef{\obs}}{\beldef{\param}} \;
\probcond{\beldef{\param}}{\excludedef{\Obs^{\cluster, \leq \timecurr}}}}{\beldef{\param}}{} \;,
\\ & \cluster \text{ existing, instantiated at } \timecurr \\[1.5ex]
\excludedef{\numcount^{\cluster, < \timecurr}}
& \integral{\probcond{\worlddef{\obs}}{\beldef{\param}} \;
\brackets{\integral{\survivegen(\beldefprev{\attr}) \;
\pdftransgen{\beldef{\param}}{\beldefprev{\param}} \;
\probcond{\beldefprev{\param}}{\excludedef{\Obs^{\cluster, < \timecurr}}}}
{\beldefprev{\param}}{}}}{\beldef{\param}}{} \;,
\\ & \cluster \text{ existing, not instantiated at } \timecurr \\[1.5ex]
\dpconc
& \integral{\probcond{\worlddef{\obs}}{\beldef{\param}} \; \pdfbase{\beldef{\param}}}{\beldef{\param}}{} \;,
\\ & \cluster \text{ new}
\end{cases}
\end{align}

We can now substitute the expressions for $\probcond{\worlddef{\obs}}{\beldef{\param}}$, $\transgen$, and $\dpbase$,
where properties of the normal distribution will help us evaluate the integrals.
The derivations in Appendix~\ref{sec.appendix-worldmodel} give the following expressions,
as well as details for finding the posterior hyperparameters $\belpmf$, $\beldef{\belmean}$, and $\beldef{\belcov}$
(recall $\beldef{\param} = (\beldeffixed{\attr}, \beldef{\pose}),
\worlddef{\obs} = (\worlddef{\attrobs}, \worlddef{\poseobs}))$:
\begin{align}
\label{eqn.forward-worldmodel}
\PP &\parensmid{\worlddef{\assn} = \cluster}{\worlddef{\obs}, \excludedef{\Obs^{\leq \timecurr}}, \excludedef{\Assn^{\leq \timecurr}}}
\nonumber \\
&\propto \begin{cases}
\excludedef{\numcount^{\cluster, \leq \timecurr}}
& \brackets{\sum_{\beldeffixed{\attr}} \attrnoisefn{\beldeffixed{\attr}}{\worlddef{\attrobs}} \;
\belpmf(\beldeffixed{\attr})} \;
\pdfnormal{\worlddef{\poseobs}}{\beldef{\belmean}, \beldef{\belcov} + \sensecov} \;,
\\ & \cluster \text{ existing, instantiated at } \timecurr \\[1.5ex]
\survivegen(\beldeffixed{\hat{\attr}}) \; \excludedef{\numcount^{\cluster, < \timecurr}}
& \brackets{\sum_{\beldeffixed{\attr}} \attrnoisefn{\beldeffixed{\attr}}{\worlddef{\attrobs}} \;
\belpmf(\beldeffixed{\attr})} \;
\pdfnormal{\worlddef{\poseobs}}{\beldefprev{\belmean},
\beldefprev{\belcov} + (\timecurr - \timeprev) \transcov(\beldeffixed{\hat{\attr}}) + \sensecov} \;,
\\ & \cluster \text{ existing, not instantiated at } \timecurr \\[1.5ex]
\dpconc
& \brackets{\sum_{\beldeffixed{\attr}} \attrnoisefn{\beldeffixed{\attr}}{\worlddef{\attrobs}} \;
\attrprior(\beldeffixed{\attr})} \;
\unif(\volume(\mathrm{world})) \;,
\\ & \cluster \text{ new}
\end{cases}
\end{align}
In the second case, for tractability in filtering, we have assumed that a cluster's dynamics
behaves according to its most-likely type $\beldeffixed{\hat{\attr}}$;
otherwise, the posterior is a mixture of Gaussians (over all possible transition densities).
Also, the third case contains an approximation to avoid evaluating an improper probability density;
see Appendix~\ref{sec.appendix-worldmodel} for details.

\section{Incorporating World Modeling Constraints}
\label{sec.constraints}

So far, we have only applied a generic DDPMM to our observations,
but have ignored the cannot-link constraint,
as well as false positives and negatives.
We now present modifications to the Gibbs sampler to handle these constraints;
the modifications are similar to those from the static case in \citet{Wong2015}
\todo{in the previous chapter (see Section~\ref{sec.todo})}.

The cannot-link constraint (see Section~\ref{sec.clc};
\review{referred to as ``one measurement per object'' (OMPO) in \citet{Wong2015} \todo{the previous chapter}})
couples together cluster assignments for observations within the same view,
since we must ensure that no two observations can be assigned to the same existing cluster.
For each view, all cluster assignments must be considered together as a
joint correspondence vector, and the probability of choosing one such correspondence
is proportional to the product of the individual cluster assignment probabilities
given in Equation~\ref{eqn.forward-worldmodel}.
Invalid correspondence vectors that violate the cannot-link constraint
are assigned zero probability and hence are not considered;
the remaining conditional probabilities are normalized.
This can be interpreted as performing \emph{blocked} Gibbs sampling,
where blocks are determined by the joint constraints:
\begin{align}
\probcond{\worlddefview{\assnview}}{\worlddefview{\obsview}, \excludedefview{\Obs}, \excludedefview{\Assn}}
&\propto \brackets{\prod_\defind
\probcond{\worlddefviewind{\assn}}{\worlddefviewind{\obs}, \excludedefview{\Obs}, \excludedefview{\Assn}}} \;
\II \brackets{\worlddefview{\assnview} \text{ satisfies CLC}}
\end{align}
The \emph{correspondence vector} $\worlddefview{\assnview}$ is again the concatenation
of the individual $\worlddefviewind{\assn}$ assignment variables,
for all observation indices $\defind$ made in view $\view$ at epoch $\timecurr$;
the interpretation of $\worlddefview{\obsview}$ is similar.
The individual terms in the product are given by Equation~\ref{eqn.forward-worldmodel}
(with the appropriate case depending on the value of $\worlddefviewind{\assn}$),
except now all observations within the same view are excluded
(since their assignments are being sampled together) --
$\excludedefview{\Obs}$ instead of $\excludedef{\Obs}$,
and likewise for assignments $\excludedefview{\Assn}$ and counts $\excludedefview{\numcount}$.
\todo{Mention exchangeability.}


For false positives, we essentially treat it as a special ``cluster'' that has no underlying parameter
Instead, we assume that if an observation is generated from a false positive,
it is generated from some spurious parameter drawn from the base distribution $\dpbase$,
so the likelihood term is the same as that for drawing a new cluster.
Like the other cases, we also multiply the likelihood by the number of points
already assigned to the cluster, i.e., the number of false positives except for those in the current view.
If there are currently no other false positives, then we multiply by the concentration parameter $\dpconc$ instead
to ensure that it is always feasible to assign observations to the false positive ``cluster''.
Also, to incorporate the assumption that false positives are generated with a fixed rate $\pfp$,
we attach a Bernoulli probability to each case in the Gibbs sampler.
The false positive conditional probability is multiplied by $\pfp$,
and all other cases are multiplied by $\parens{1-\pfp}$.
In summary, the conditional probability of an observation being a false positive ($\assn = 0$) is:
\todo{Put $\pfp$ outside?}
\begin{align}
\label{eqn.falsepositive}
\probcond{\worlddefviewind{\assn} = 0}{\worlddefviewind{\obs}, \excludedefview{\Obs}, \excludedefview{\Assn}}
\propto \brackets{\sum_{\beldeffixed{\attr}} \attrnoisefn{\beldeffixed{\attr}}{\worlddefviewind{\attrobs}} \;
\attrprior(\beldeffixed{\attr})} \;
\unif(\volume(\mathrm{world})) \times
\begin{cases}
\pfp \excludedefview{\numcount}^0 \;,
& \excludedefview{\numcount}^0 > 0 \\
\pfp \dpconc \;,
& \excludedefview{\numcount}^0 = 0
\end{cases}
\end{align}
The normalizer depends on the other cases in Equation~\ref{eqn.forward-worldmodel}
(with additional $\parens{1-\pfp}$ factors).

Finally, for false negatives, recall that an object that is within the field of view
fails to be detected with type-dependent probability $\pfn(\beldeffixed{\attr})$.
Let $\worlddefview{\delta}_\cluster$ be $1$ if cluster $\cluster$ is detected
in view $\view$ at epoch $\timecurr$, and $0$ otherwise.
For a cluster $\cluster$ that is alive at epoch $\timecurr$
($\beldeffixed{\timebirth} \leq \timecurr \leq \beldeffixed{\timedeath}$)
with parameter $\beldef{\param}$, the probability of detection is therefore:
\begin{align}
\label{eqn.detection}
\prob \parens{\worlddefview{\delta}_\cluster = 1} = \brackets{1 - \pfn(\beldeffixed{\attr})} \; \prob \parens{\beldef{\param} \in \reg}
= \brackets{1 -\sum_{\beldeffixed{\attr}} \pfn(\beldeffixed{\attr}) \; \belpmf(\beldeffixed{\attr})} \;
\pdferfgen{\beldef{\pose} \in \reg}{\beldef{\belmean}, \beldef{\belcov}}
\end{align}
The $\erfgen$ function denotes the CDF of the multivariate normal distribution,
with mean $\beldef{\belmean}$ and covariance $\beldef{\belcov}$. \todo{In practice.}
For a particular view $\reg$, we only evaluate the above detection probability
on clusters that are currently alive at epoch $\timecurr$.
For each such cluster, there is a corresponding $\worlddefview{\delta}_\cluster$ detection indicator variable,
whose value is determined during sampling by the candidate joint correspondence vector $\worlddefview{\assnview}$:
if some element of $\worlddefview{\assnview}$ is assigned to cluster index $\cluster$,
then $\worlddefview{\delta}_\cluster = 1$; otherwise, $\worlddefview{\delta}_\cluster = 0$.
The detection probability for the correspondence vector is:
\todo{What about revived and new clusters? No FN evaluation? Intermediate time steps? Write out likelihood.}
\begin{align}
\prob_{\text{D}} \parensmid{\worlddefview{\assnview}}{\excludedefview{\Obs}, \excludedefview{\Assn}}
= \prod_{\cluster: \; \beldeffixed{\timebirth} \leq \timecurr \leq \beldeffixed{\timedeath}}
\brackets{\prob \parens{\worlddefview{\delta}_\cluster = 1}}^{\worlddefview{\delta}_\cluster}
\brackets{1-\prob \parens{\worlddefview{\delta}_\cluster = 1}}^{1-\worlddefview{\delta}_\cluster}
\end{align}

\todo{Write out the algorithm?}
Putting everything together, we arrive at a constrained blocked collapsed Gibbs sampling inference algorithm.
The algorithm takes the observations $\Obs = \braces{\worlddefviewind{\obs}}$ and visible regions $\braces{\reg}$ as input.
As output, the algorithm produces samples from the posterior distribution over correspondence vectors
$\braces{\worlddefview{\assnview}}$, from which we can compute the posterior parameter distributions
$\beldeffixed{\attr} \sim \belpmf$ and $\beldef{\pose} \sim \normal \parens{\beldef{\belmean}, \beldef{\belcov}}$.
The sampling algorithm repeatedly iterates over epochs $\timecurr$ and views $\view$,
each time sampling a new correspondence vector $\worlddefview{\assnview}$ from its constrained conditional distribution:
\begin{align}
\label{eqn.gibbs-view}
\prob_{\text{View}} \parensmid{\worlddefview{\assnview}}{\worlddefview{\obsview}, \excludedefview{\Obs}, \excludedefview{\Assn}}
&\propto \brackets{\prod_{\defind: \; \worlddefviewind{\assn} \neq 0} \parens{1-\pfp} \;
\probcond{\worlddefviewind{\assn}}{\worlddefviewind{\obs}, \excludedefview{\Obs}, \excludedefview{\Assn}}} \nonumber \\
&\times \brackets{\prod_{\defind: \; \worlddefviewind{\assn} = 0} \pfp \;
\brackets{\sum_{\beldeffixed{\attr}} \attrnoisefn{\beldeffixed{\attr}}{\worlddefviewind{\attrobs}} \;
\attrprior(\beldeffixed{\attr})} \;
\frac{1}{\volume(\mathrm{world})} \times
\begin{cases}
\excludedefview{\numcount}^0 \;,
& \excludedefview{\numcount}^0 > 0 \\
\dpconc \;,
& \excludedefview{\numcount}^0 = 0
\end{cases}} \nonumber \\
&\times \brackets{\prod_{\cluster: \; \beldeffixed{\timebirth} \leq \timecurr \leq \beldeffixed{\timedeath}}
\brackets{\prob \parens{\worlddefview{\delta}_\cluster = 1}}^{\worlddefview{\delta}_\cluster}
\brackets{1-\prob \parens{\worlddefview{\delta}_\cluster = 1}}^{1-\worlddefview{\delta}_\cluster}} \nonumber \\
&\times \II \brackets{\worlddefview{\assnview} \text{ satisfies CLC}}
\end{align}
The probability terms in the first and third lines can be found in
Equations~\ref{eqn.forward-worldmodel} and \ref{eqn.detection} respectively.

\review{As in the static case, after incorporating the world modeling constraints,
inference becomes inefficient because we now have to compute conditional probabilities for
(and sample from) the joint space of correspondence vectors,
which in general is exponential in the number of observations in a view.
Using the same insights and ideas as \review{in \citet{Wong2015}} \todo{before}, however,
we can adaptively factor the correspondence vector by initially decoupling all assignment variables,
then coupling only those that violate the cannot-link constraint
\todo{; see Section~\ref{sec.todo} for details}.}

\iffactored
\subsection{An online method to incorporate view-level constraints}
\fi

\section{Approximate Maximum \textit{a Posteriori} (MAP) inference}

We have now presented the entire Gibbs sampling algorithm for DDPMM-based world modeling.
However, sampling-based inference can be slow, especially because of the cannot-link constraint
that couples together many latent variables, even if adaptive factoring is used.
Although we are interested in maintaining an estimate of our uncertainty in the world,
\review{frequently just having the most-likely (maximum \textit{a posteriori} -- MAP) world state suffices}.
In general, even the MAP world model is hard to find \citep{BarShalom1988},
and many approximate solutions have been proposed.
\todo{Specific citation for hardness? See Robert Collins slides as well.
Multidimensional ($T \geq 3$) assignment problem is NP-complete (Karp).}

In the static case, \citet{Wong2015} \todo{we} adapted a hard-clustering algorithm, DP-means,
and empirically found that it returned good clustering assignments for some hyperparameter settings
\todo{(see Algorithm~\ref{fig.todo})}.
A similar analysis via small-variance asymptotics was performed recently for DDPs,
where the mixture components were Gaussian distributions with isotropic noise,
resulting in the Dynamic Means algorithm \citep{Campbell2013}.
However, there is no simple and principled way to incorporate the additional information from
Section~\ref{sec.constraints}. Additionally, even without such modifications,
the Dynamic Means algorithm requires three free hyperparameters to be specified,
which may be significantly harder to tune than the one in DP-means.
Instead, we will use a much older idea that does not involve asymptotics,
can incorporate all the world-modeling information and constraints,
and produces an local optimization algorithm that is similar in spirit to Dynamic Means.

\subsection{Iterated conditional modes (ICM)}

The \textit{iterated conditional modes} (ICM) algorithm performs coordinate ascent
on each variable's conditional distribution, and is guaranteed to converge to a local maximum \citep{Besag1986}.
In particular, instead of iteratively sampling correspondence vectors
from their conditional distributions in Gibbs sampling, we find the most-likely one,
update parameters based on it, and repeat for each view.
Since we are still dealing with the joint space of assignments for all observations in a given view,
finding the maximizer still potentially requires searching through a combinatorial space.
Fortunately, finding the most-likely correspondence can be formulated as a
maximum weighted assignment problem, for which efficient algorithms such as the Hungarian algorithm exist
(and have been previously used in data association).
\todo{ICM idea is very similar to Collins!!!}

Suppose, for view $\view$ at epoch $\timecurr$, there are $\numobs$ observations
$\braces{\obs_1, \ldots, \obs_\numobs}$ and $\numcluster$ existing clusters
\review{(possibly not alive/instantiated)}.
Then we wish to match each $\obs_i$ to an existing cluster,
a new cluster, or a false positive.
Any unmatched existing cluster must also be assigned
the probability of missed detection.
We can solve this as an assignment problem with the following payoff matrix:
\begin{center}
\begin{tabular}{r|c|c|}
& Obs ($\numobs$) & FN ($\numobs+\numcluster$) \\
\hline
Clusters ($\numcluster$) & $\log \prob \parens{\assn_i = \cluster} +
\log (1 - \pfp)$ &
$\II \brackets{\beldeffixed{\timebirth} \leq \timecurr \leq \beldeffixed{\timedeath}} \; \log \prob \parens{\delta_\cluster = 0}$ \\
\todo{Vert align} & $+ \; \II \brackets{\beldeffixed{\timebirth} \leq \timecurr \leq \beldeffixed{\timedeath}} \;
\log \prob \parens{\delta_\cluster = 1}$ & \\
\hline
New ($\numobs$) & $\log \prob \parens{\assn_i = \text{new}} +
\log (1 - \pfp)$ & 0 \todo{Preferred if $\dpconc (1-\pfp) \geq \pfp N^0$ (or $\dpconc$)} \\
\hline
FP ($\numobs$) & $\log \prob \parens{\assn_i = 0 \; (\text{FP})} + \log \pfp$ & 0 \\
\hline
\end{tabular}
\end{center}

The payoff matrix has $2\numobs+\numcluster$ entries (indicated in parentheses),
to allow for the case that all observations are assigned to new clusters,
and likewise that all are spurious.
Any extra New/FP nodes are assigned to extra FN nodes, with zero payoff.
The payoffs in the first column are:
for an existing cluster,
given by cases 1 and 2 in Equation~\ref{eqn.forward-worldmodel},
depending on whether or not the cluster has been instantiated yet;
for a new cluster, given by case 3 in Equation~\ref{eqn.forward-worldmodel};
and for a false positive, given by Equation~\ref{eqn.falsepositive}.
Note that log probabilities are used to decompose the view's
joint correspondence probability into a sum of individual terms.
By construction, the cannot-link constraint is satisfied.

\todo{Show correctness. Complexity? State some theorems? Murty's k-best?}

\subsection{A two-stage inference scheme}

\begin{figure}[t!]
\small
  \centering
  \begin{subfigure}{.48\textwidth}
    \centering
    \begin{algorithmic}[1]
      \REQUIRE Observations $\Obs = \braces{\worlddefviewind{\obs}}$ \\ Visible regions $\braces{\reg}$ \\
      Number of samples $\numcount$
      \ENSURE Samples of cluster assignments $\braces{\worlddefview{\assnview}}$
      \STATE Init. all entries to $-1$ (FP) in $\Assn^{(0)} = \braces{\worlddefview{\assnview}}^{(0)}$
      \FOR {$n := 1$ \TO $\numcount$}
        \STATE $\Assn' := \mathrm{Proposal}(\Assn^{(n-1)})$ (see \citet{Oh2009})
        \STATE $A \parens{\Assn^{(n-1)} \rightarrow \Assn'} :=$ \\ $\min \parens{1, \frac{\mathrm{Likelihood}(\Assn') \; \PP \parens{\Assn' \rightarrow \Assn^{(n-1)}}}{\mathrm{Likelihood}(\Assn^{(n-1)}) \; \PP \parens{\Assn^{(n-1)} \rightarrow \Assn'}}}$
        \STATE Sample $u \sim \unif(0,1)$
        \IF {$u < A \parens{\Assn^{(n-1)} \rightarrow \Assn'}$}
          \STATE $\Assn^n = \Assn'$
        \ELSE
          \STATE $\Assn^n = \Assn^{(n-1)}$
        \ENDIF
      \ENDFOR
    \end{algorithmic}
    \caption{MCMCDA \citep{Oh2009} for DDPMM}
    \label{alg.ddpmm.mcmcda}
  \end{subfigure}
  \hfill
  \begin{subfigure}{.48\textwidth}
    \centering
    \begin{algorithmic}[1]
      \REQUIRE Observations $\Obs = \braces{\worlddefviewind{\obs}}$ \\ Visible regions $\braces{\reg}$ \\
      Number of samples $\numcount$
      \ENSURE Samples of cluster assignments $\braces{\worlddefview{\assnview}}$
      \STATE Init. all entries to $-1$ (FP) in $\Assn^{(0)} = \braces{\worlddefview{\assnview}}^{(0)}$
      \REPEAT
        \FOR {$\timecurr := 1$ \TO $T$; $\view := 1$ \TO $V^\timecurr$}
          \STATE Solve ICM weighted assignment problem \\ for most-likely $\worlddefview{\assnview}$,
          given $\exclude{\Assn^\timecurr}{\view}$
        \ENDFOR
      \UNTIL{convergence}
      \STATE \review{Construct new dataset $C = \braces{c^\timecurr_\defind}$ with \\
      a single data point for each non-FP cluster \\ found by ICM (at the same epoch)}
      \STATE \review{Sample tracks $L$ by performing MCMCDA on $C$}
      \STATE \review{Convert track samples to cluster assignments}
    \end{algorithmic}
    \caption{Two-stage inference algorithm for DDPMM}
    \label{alg.ddpmm.mcmc_icm}
  \end{subfigure}
  \caption[Inference algorithms for dependent Dirichlet-process mixture models (DDPMMs).]
  {Two algorithms for performing inference in DDPMMs, one by Metropolis-Hastings (MH) \citep{Oh2009}, the other a two-stage procedure involving ICM, followed by the MH procedure.}
  \label{alg.ddpmm}
\end{figure}

\todo{The inference algorithms we have discussed so far,
and other traditional tracking algorithms such as MHT,
all consider each view in sequence, sampling/scoring correspondence vectors
given the associations from all previous views.
Although the Gibbs sampler will eventually converge to the true posterior distribution,
convergence tends to be slow in practice because it favors local optima.}

\todo{MCMC methods such as MCMCDA can consider tracks across multiple times (views/epochs),
but applying it directly on our problem gives poor results because of how likelihoods are scored.
The feasible neighborhood is large, but only very few neighbors give high-scoring associations.}

Although the ICM algorithm presented can find good clusters at a single epoch very quickly,
we will see in experiments that it does not converge to good cluster trajectories.
The issue is that ICM moves are local, in that it considers one view at a time.
Suppose we have identified correctly all objects in epoch $1$ using ICM.
When we consider the first view in epoch $2$, there may be significant changes present,
and using the first view only, ICM decides whether or not to assign the new observations
to existing clusters (by reviving them from the previous epoch).
Since the uncertainty in the object states immediately after a transition is high,
basing the cluster connectivity decisions on a single view is unreliable.
\todo{Show a concrete example.}

This suggests a two-level inference scheme. Since ICM can reliably find good clusters within single epochs,
we first apply ICM to each epoch's data \emph{independently}, treating them as unrelated static worlds.
Next, we attempt to connect clusters between different epochs.
This is essentially another tracking problem, although the likelihood function is somewhat different
(depends on many underlying data points), and is much reduced in size.
Since the problem is significantly smaller, traditional tracking methods such as MHT can be applied
to this cluster-level tracking problem.

We present one such scheme in Algorithm~\ref{alg.ddpmm.mcmc_icm},
using MCMCDA (Algorithm~\ref{alg.ddpmm.mcmcda}; \citet{Oh2009}) to solve the cluster-level problem.
We choose a batch-mode sampling algorithm such as MCMCDA
because it can return samples from the posterior distribution,
and has an attractive anytime property --
we can terminate at any point and still return a list of valid samples.
For inferring the MAP configuration, the best sample can be returned instead.
Since we are sampling from the true posterior distribution,
in the limit of infinite samples, the true MAP configuration will be found almost surely.
\todo{Talk about FPs.}

To apply MCMCDA, we need to evaluate the likelihood of a complete configuration $\Assn$,
encompassing all epochs and views (line 4 in Algorithm~\ref{alg.ddpmm.mcmcda}).
To do so, we first find the posterior parameter distributions
for the clusters/objects (as given by $\Assn$) using Appendix~\ref{sec.appendix-worldmodel},
then combine the observation likelihoods (\review{Equation~\ref{eqn.marginal-likelihood}}),
as well as the false positive and false negative priors: \todo{Assume $\Assn$ satisfies CLC.}
\todo{FP, FN should belong to $\prob(\Assn)$? Birth/survive/death?
Finding $\worlddefview{\delta}_\cluster$ only requires examining $\worlddefview{\assnview}$,
but evaluating $\prob \parens{\worlddefview{\delta}_\cluster = 1}$
also needs parameter posterior (and ranges).}
\begin{align}
\label{eqn.score}
\probcond{\Obs}{\Assn} \prob \parens{\Assn} = \prod_\timecurr \prod_\view \;
&\probcond{\worlddefview{\obsview}}{\worlddefview{\assnview}} \;
\prob_{\mathrm{FP}} \parens{\worlddefview{\assnview}} \;
\prob_{\mathrm{FN}} \parens{\worlddefview{\assnview}} \nonumber \\
= \prod_\timecurr \prod_\view \; &\Bigg\{
\brackets{\prod_\defind \integral{\probcond{\worlddef{\obs}}{\beldef{\param}, \worlddef{\assn} = \cluster} \;
\prob \parens{\beldef{\param}}}{\beldef{\param}}{}} \nonumber \\
&\times \bin \parensmid{\worlddefview{\numcount}_{\assn = 0}}{\worlddefview{\numcount}, \pfp} \nonumber \\
&\times \brackets{\prod_{\cluster: \; \beldeffixed{\timebirth} \leq \timecurr \leq \beldeffixed{\timedeath}}
\brackets{\prob \parens{\worlddefview{\delta}_\cluster = 1}}^{\worlddefview{\delta}_\cluster}
\brackets{1-\prob \parens{\worlddefview{\delta}_\cluster = 1}}^{1-\worlddefview{\delta}_\cluster}} \Bigg\}
\end{align}

\section{Experiments}

Approximate MAP inference for world modeling via
ICM, MCMCDA, and the two-stage algorithm ICM-MCMC were tested on a simulated domain,
and also on a sequence of real robot vision data constructed
from the static scenes in \citet{Wong2015} \todo{the previous chapter}.
To perform MAP inference on MCMCDA and ICM-MCMC, the most-likely sample
(as scored by Equation~\ref{eqn.score}) was chosen,
from $10^5$ samples in MCMCDA, and $10^4$ in the second stage of ICM-MCMC.
In both experiments, ICM-MCMC significantly outperforms the other two methods,
and even ICM performs better than MCMCDA.

\begin{figure}[t]
\centering
  \begin{subfigure}{\textwidth}
    \centering
    \begin{subfigure}{.45\textwidth}
      \centering
      \includegraphics[width=\textwidth]{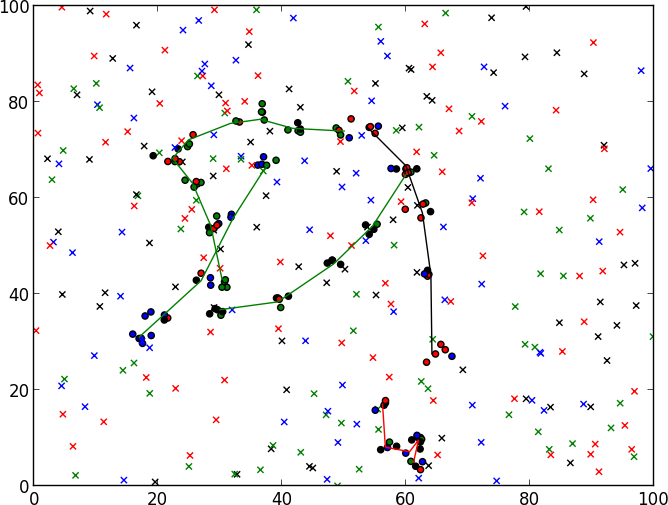}
      \caption{True object trajectories}
    \end{subfigure}
    \qquad
    \begin{subfigure}{.45\textwidth}
      \centering
      \includegraphics[width=\textwidth]{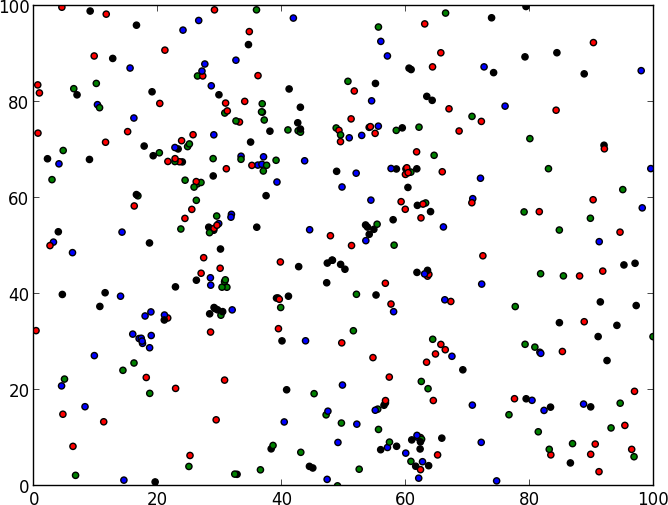}
      \caption{Data (from all $10$ epochs)}
    \end{subfigure}
  \end{subfigure}\\
  \begin{subfigure}{\textwidth}
    \centering
    \begin{subfigure}{.24\textwidth}
      \centering
      \includegraphics[width=\textwidth]{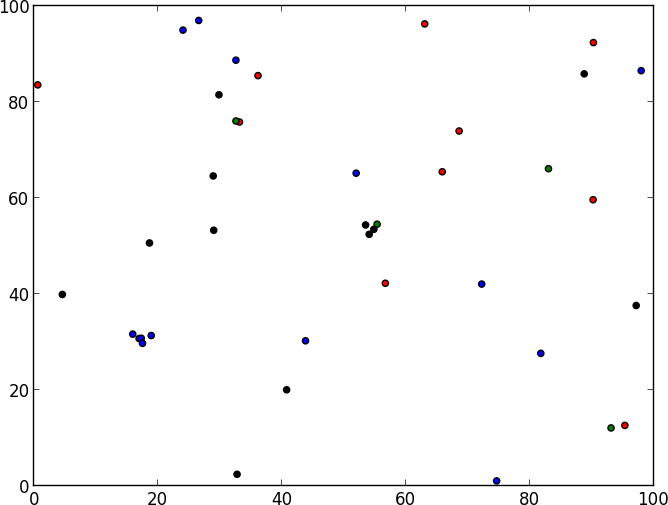}
      \caption{$\timecurr = 5$}
    \end{subfigure}
    \hfill
    \begin{subfigure}{.24\textwidth}
      \centering
      \includegraphics[width=\textwidth]{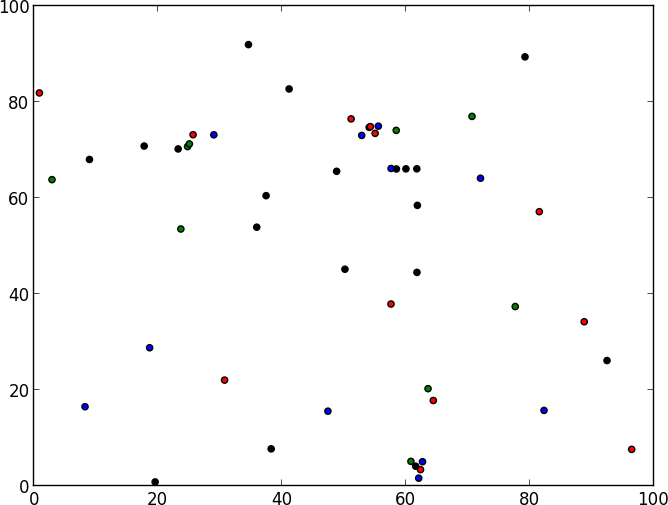}
      \caption{$\timecurr = 6$}
    \end{subfigure}
    \hfill
    \begin{subfigure}{.24\textwidth}
      \centering
      \includegraphics[width=\textwidth]{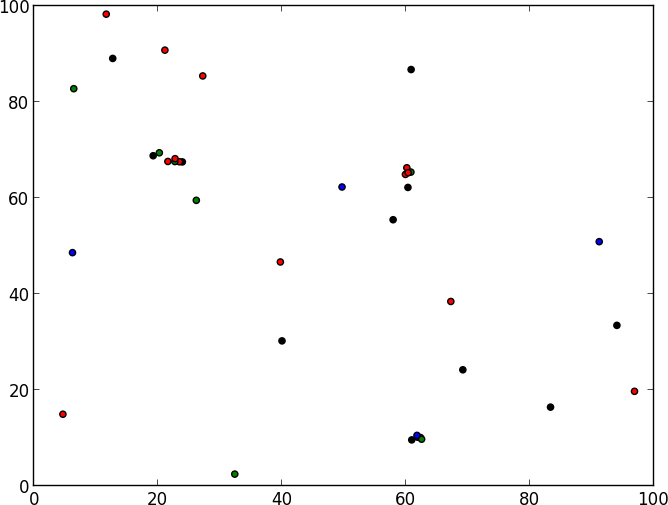}
      \caption{$\timecurr = 7$}
    \end{subfigure}
    \hfill
    \begin{subfigure}{.24\textwidth}
      \centering
      \includegraphics[width=\textwidth]{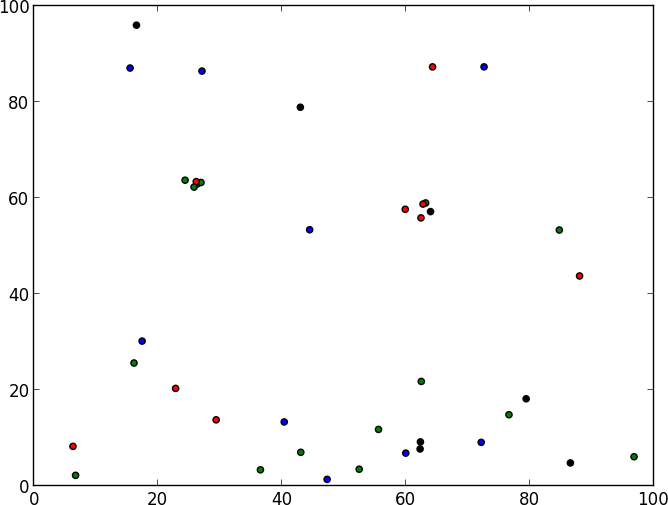}
      \caption{$\timecurr = 8$}
    \end{subfigure}
  \end{subfigure}
\caption[Data and true object states in a simulated domain.]
{Data and object states in a simulated domain.
The top left shows the true object $(x,y)$ locations and their trajectories over time,
color-coded by their associated object type.
Observations are shown as filled dots (corresponding to true positives)
and crosses (false positives).
The top right shows the data from all $10$ epochs ($5$ views per epoch) that is given as input,
without any information about the underlying object states and associations.
Some form of clustering over views and time is visible.
A more realistic view of the data is shown in the bottom row,
for a sequence of $4$ epochs.}
\label{fig.sim_data}
\end{figure}

\subsection{Simulation}

Objects in our simulated domain had one of four fixed object types,
a time-evolving location $(x,y) \in [0,100] \times [0,100]$,
and a time-evolving velocity vector.
Observations were made in $10$ epochs of this domain,
with $5$ views per epoch (visible region is the entire domain).
In total, $5$ objects existed, each for some contiguous sub-interval of the elapsed time.
Within each view, the number of false positives was generated from $\poi(5)$,
and the probability of a missed detection was $0.1$.
The correct object type was observed with probability $0.6$,
with equal likelihood ($0.1$) of being confused with the other $3$ object types.
Locations were observed with isotropic Gaussian noise, standard deviation $1.0$.
The object's velocity vector was maintained from the previous time step,
with added Gaussian noise, standard deviation $5.0$.
Between epochs, the probability of survival was $0.9$.
The observed data (i.e., the algorithm input)
and the true object states are shown in Figure~\ref{fig.sim_data}.

\begin{figure}
\centering
  \begin{subfigure}{\textwidth}
    \centering
    \begin{subfigure}{.45\textwidth}
      \centering
      \includegraphics[width=\textwidth]{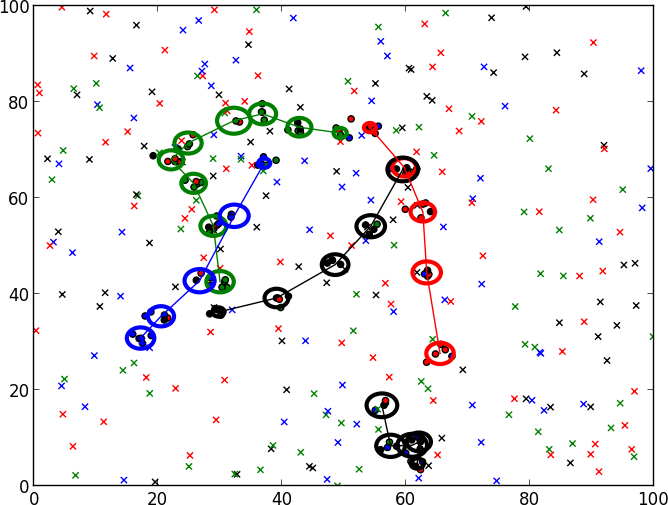}
      \caption{Truth}
    \end{subfigure}
    \qquad
    \begin{subfigure}{.45\textwidth}
      \centering
      \includegraphics[width=\textwidth]{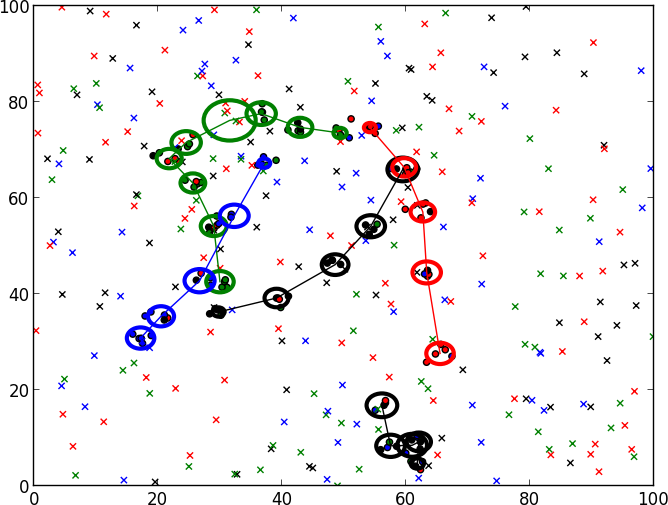}
      \caption{ICM-MCMC (LL $= -3168$)}
    \end{subfigure}
  \end{subfigure}\\
  \begin{subfigure}{\textwidth}
    \centering
    \begin{subfigure}{.45\textwidth}
      \centering
      \includegraphics[width=\textwidth]{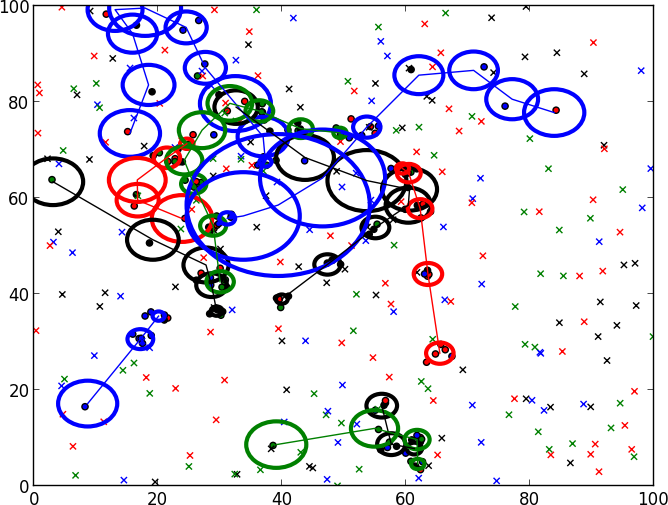}
      \caption{ICM (LL $= -3329$)}
    \end{subfigure}
    \qquad
    \begin{subfigure}{.45\textwidth}
      \centering
      \includegraphics[width=\textwidth]{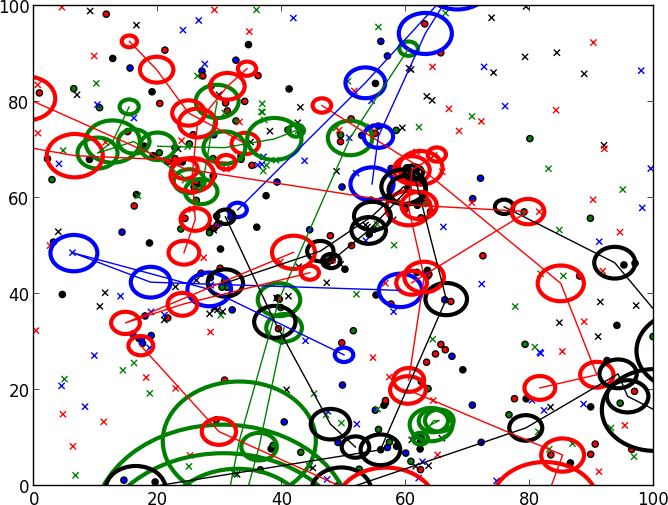}
      \caption{MCMC (LL $= -3365$)}
    \end{subfigure}
  \end{subfigure}
\caption[Cluster trajectories in the simulated domain, found using three different algorithms.]
{The clusters found for the simulated domain are shown in thick ellipses,
centered at the location mean, color-coded by the most-likely object type inferred
(across the entire trajectory, since it is a static attribute).
The ellipses depict a level set of the posterior location distribution
(uncertainty given by Gaussian covariance matrix).
The posterior clusters derived from the true association is shown in the top left;
the one found by ICM-MCMC is essentially identical (with a minor difference in the green track).
In contrast, the posterior clusters found by ICM and the most-likely sample from MCMC (of $10^5$),
shown in the bottom row, are qualitatively much different,
and have significantly lower log-likelihood (LL) values.}
\label{fig.sim_tracks}
\end{figure}

The resulting MAP clusters found by ICM, MCMCDA, and ICM-MCMC are shown in Figure~\ref{fig.sim_tracks},
along with their log-likelihood values (higher / less negative is better).
ICM-MCMC clearly outperforms the other methods, and finds essentially the same clusters
as given by the true association. The clusters found generally have tight covariance values,
unlike those in ICM and MCMCDA. These two methods, especially MCMCDA,
tend to find many more clusters than are truly present.

\subsection{Using robot data from static scenes}

\begin{figure}
\centering
  \begin{subfigure}{\textwidth}
    \centering
    \begin{subfigure}{.24\textwidth}
      \centering
      \includegraphics[height=.4\textheight]{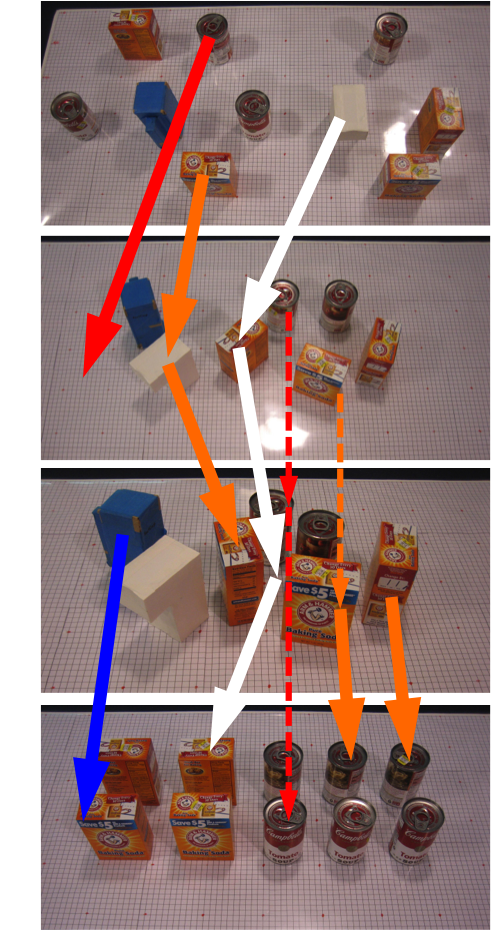}
      \caption{ICM transitions not present in ICM-MCMC}
    \end{subfigure}
    \begin{subfigure}{.24\textwidth}
      \centering
      \includegraphics[height=.4\textheight]{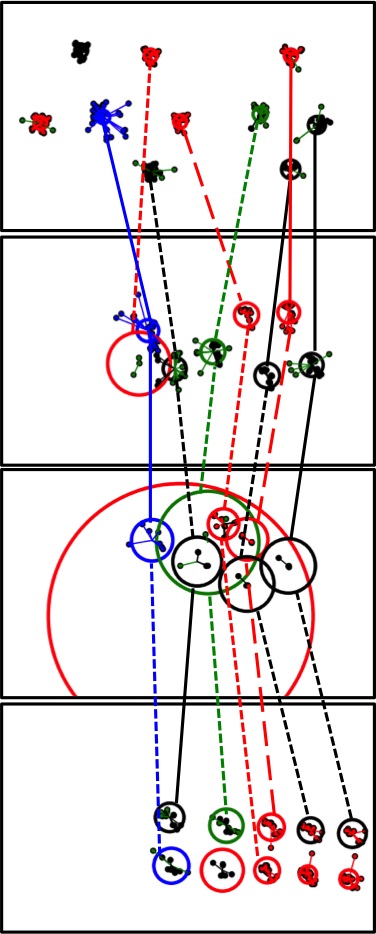}
      \caption{Most-likely ICM configuration (LL $= -968$)}
    \end{subfigure}
    \begin{subfigure}{.24\textwidth}
      \centering
      \includegraphics[height=.4\textheight]{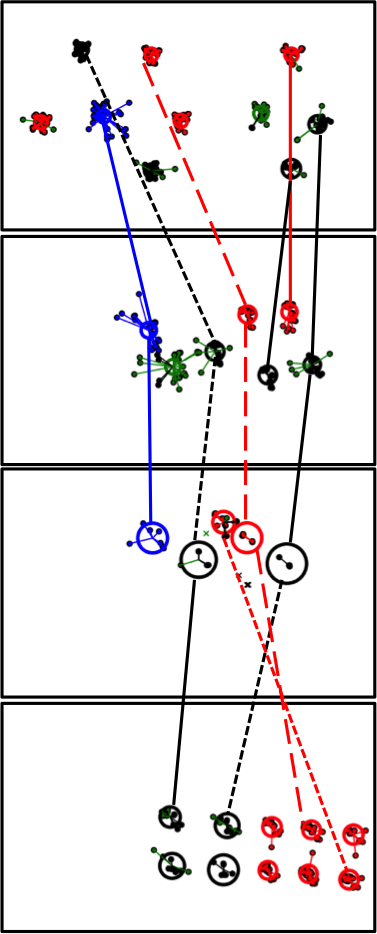}
      \caption{Most-likely ICM-MCMC conf (LL $= -931$)}
    \end{subfigure}
    \begin{subfigure}{.24\textwidth}
      \centering
      \includegraphics[height=.4\textheight]{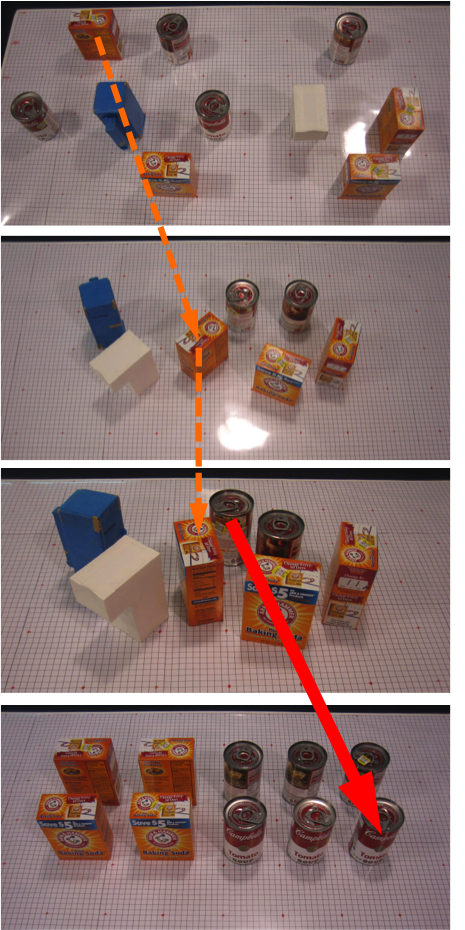}
      \caption{ICM-MCMC transitions not present in ICM}
    \end{subfigure}
  \end{subfigure}
\caption[Inferring object trajectories in robot vision data from static scenes]
{Approximate MAP cluster (object) trajectories found using ICM and ICM-MCMC
on the robot vision data collection in \citet{Wong2015} \todo{the previous chapter}.
The concatenated sequence of scenes (epochs) is shown from top to bottom.
The inferred clusters and tracks are shown in the middle two columns.
Lines connecting cluster pairs between epochs are color-coded by the inferred object type
(fixed across epochs), and are marked by one of three line styles used to
compare results from the two algorithms.
A solid line means the same pair was connected by both algorithms;
a dashed line means a similar pair (in likelihood) was connected;
a dotted line means the pair was not connected by the other algorithm.
To make the differences clearer, the top-down reference views
have been annotated with arrows, for pairs of objects that were
only connected by one algorithm (dotted lines in the middle two).
The left column shows pairs that were connected by ICM but not ICM-MCMC;
the right column shows the opposite.
Solid arrows depict transitions that are unlikely,
whereas dashed arrows depict plausible transitions.
ICM tends to suggest many more transitions than ICM-MCMC,
many of which are actually implausible.}
\label{fig.swm}
\end{figure}

We also applied the same algorithms to the static robot vision data
that were used in \citet{Wong2015} \todo{the previous chapter} to evaluate DPMM methods.
To convert static scenes into dynamic scenes,
we choose static scenes that were reasonably similar,
and simply concatenated their data together,
as if each scene corresponded to a different epoch.
One such example is shown in Figure~\ref{fig.swm}.

Objects in different scenes were all placed on the same tabletop of dimensions
$1.2\mathrm{m} \times 0.6\mathrm{m}$;
all data were placed in the table's frame of reference.
Four object types were present, and typically each scene had $5$--$10$ objects.
Unlike the previous simulation, we do not assume objects have velocities;
between epochs, we assume that the location changes with isotropic Gaussian noise,
standard deviation $0.1$.
Since changes were significant between epochs, we assumed a relatively low $0.5$
probability of survival.
Object locations are sensed with Gaussian noise, standard deviation $0.03$;
the object type noise model and probability of detection is the same as before.
The probability of false positives is much lower for this domain;
we assumed the number of false positives had a $\poi(0.1)$ distribution.

Figure~\ref{fig.swm} shows the MAP associations found by ICM and ICM-MCMC,
with lines connecting cluster states over epochs.
Annotations were also added (in the form of three different line styles)
to facilitate comparison between the ICM and ICM-MCMC results;
\review{see figure caption for details.
ICM tends to suggest many more transitions than ICM-MCMC,
many of which are actually implausible.}

\newpage

\appendix


\section{Background on dependent Dirichlet processes}
\label{sec.appendix-ddp}

\citet{Lin2010} exploited the fact that there exists a one-to-one correspondence
between DPs over space $\Omega$ and spatial Poisson processes in the product space
$\Omega \times \mathbb{R}_+$. This means that an underlying Poisson process
can be extracted from any DP, and vice versa.
By considering transitions on the underlying Poisson processes,
\review{and restricting to transition steps where the Poisson process
remains closed under transition} (more fundamentally, by preserving complete randomness),
we obtain a new spatial Poisson process at the next time step,
which can be converted back to a new DP.

According to the stick-breaking construction of the DP \citep{Sethuraman1994},
if $D^{\timecurr} \sim \mathrm{DP}$, then it can be expressed as infinite sum of weighted atoms:
$D^{\timecurr} = \sum_{i=1}^\infty w_i \delta_{\param_i}$, where $w_i \in \mathbb{R}_+$,
and $\param_i \in \Omega$. Then the following DP-preserving transition steps
are applied in order:
\begin{itemize}
\item Subsampling (removal): Let $\survive: \Omega \rightarrow [0, 1]$
be a parameter-dependent survival rate, i.e., $\survive(\param)$ specifies
how likely some $\param$ in the current time step survives in the next time step.
For each atom $\param_i$, draw $b_i \sim \ber \parens{\survive(\param_i)}$,
and retain atoms with $b_i = 1$.
Renormalizing the weights on the retained atoms gives a new DP
$D' = \sum_{i: b_i=1} w_i' \delta_{\param_i}$ (where $\sum_{i: b_i=1} w_i' = 1$).
\item Point transition (movement): Let $\pdftrans{\cdot}{\param}: \Omega \rightarrow \mathbb{R}_+$
be a parameter-dependent transition function, i.e., $\pdftrans{\param'}{\param}$ specifies
how likely some $\param$ in the current time step moves to $\param'$ in the next time step,
given that it survives.
For each atom $\param_i$, draw $\param_i' \sim \pdftrans{\cdot}{\param}$.
Then $D'' = \sum_{i: b_i=1} w_i' \delta_{\param_i'}$ is a new DP.
\item Superposition (addition): Let $\Delta = \sum_j \varpi_j \delta_{\vartheta_j}$
be a new independent DP, and let $(c, d) \sim \mathrm{Dir} \parens{\dpconc'', \dpconc}$,
where $\dpconc''$ and $\dpconc$ are the concentration parameters of
$D''$ and $\Delta$ respectively. Then the random convex combination
$D^{\timecurr+1} = c D'' + d \Delta$ is a DP,
and acts as the prior for the next time step.
\end{itemize}

The upshot of this DDP construction is that, if we marginalize out the DP prior,
we get the following prior for $\param^{\timecurr+1}$,
given the parameters from the previous time $\Param^{\timecurr}$:
\begin{align}
\label{eqn.priorprior}
\param^{\timecurr+1} \;|\; \Param^{\timecurr} \propto \; &\dpconc \pdfbase{\param^{\timecurr+1}} + \sum_k q(\param^{\cluster \timecurr}) \; \numcount^{\cluster, \leq \timecurr} \; \pdftrans{\param^{\timecurr+1}}{\param^{\cluster \timecurr}}
\end{align}
The first term is for new atoms, drawn from a DP with base distribution $\pdfbase{\param}$
and concentration parameter $\dpconc$.\footnote{Technically, $\dpconc$ includes both the innovation process
from the superposition step, as well as a subsampled and transitioned version of innovation processes from previous times;
see \citet{Lin2012}.}
The second term corresponds corresponds to existing atoms
that have undergone subsampling and transition steps;
these steps affect the assignment probability,
as indicated by the presence of $\survive$ and $\trans$.
Additionally, $\numcount^{\cluster,\leq t}$ is the number of points that have been assigned
to cluster $\cluster$, for all time steps up to time $\timecurr$.
\review{This term is similar to that in the DP.}
Notice that if $\survive \equiv 1$ and
$\pdftrans{\cdot}{\beldef{\param}} = \delta_{\beldef{\param}}$,
then we exactly get back the predictive distribution in the DP.

Since $\param^{\timecurr+1} \sim D^{\timecurr+1}$, and $D^{\timecurr+1}$ is a DP,
we can find the predictive distribution of $\param^{\timecurr+1}$,
conditioning also on parameters $\Param^{\timecurr+1}$
that have been instantiated at time $(\timecurr+1)$:
\begin{align}
\param^{\timecurr+1} \;|\; \Param^{\timecurr} \propto \; &\dpconc \pdfbase{\param^{\timecurr+1}} + \sum_{\cluster: \numcount^{\cluster, \timecurr+1} > 0} \numcount^{\cluster, \leq \timecurr + 1} \; \II \brackets{\param^{\cluster, \timecurr+1} = \beldef{\param}} +
\sum_{\cluster: \numcount^{\cluster, \timecurr+1} = 0} \survive(\param^{\cluster \timecurr}) \; \numcount^{\cluster, \leq \timecurr} \; \pdftrans{\param^{\timecurr+1}}{\param^{\cluster \timecurr}}
\end{align}

In general, some atoms may not be observed for several time steps,
but still affect the prior (with decayed weight and dispersed parameter values).
Also, some clusters may already have been instantiated at the current time $\timecurr$,
either newly drawn from the innovation process, or transitioned from existing atoms.
The general form of the prior on $\param^\timecurr$ is:
\begin{align}
\label{eqn.prior2}
\param^\timecurr \;|\; \Param^{\leq \timecurr} \propto \; \dpconc \pdfbase{\param^\timecurr} + &\sum_{\cluster: \beldef{N} = 0} \beldef{\survive} \; \numcount^{\cluster, \leq \timeprev} \; \pdftrans{\param^\timecurr}{\beldefprev{\param}} +
\sum_{\cluster: \beldef{\numcount} > 0} \numcount^{\cluster, \leq \timecurr} \; \II \brackets{\param = \beldef{\param}}
\end{align}
The first two terms are similar to those in Equation~\ref{eqn.priorprior} above,
except the sum is only over clusters that have not been instantiated at time $\timecurr$
($\beldef{N} = 0$). These existing clusters may be `revived', but they are weighted by
accumulated subsampling and transition terms, based on the previous time
$\timeprev = \beldef{\timeprev}$ at which they were instantiated:
\begin{align}
\label{eqn.transgen2}
\beldef{\survive} &\triangleq \brackets{\survive(\beldefprev{\param})}^{\timecurr - \timeprev} \\
\pdftrans{\param^\timecurr}{\beldefprev{\param}} &\triangleq \integral{\cdots \integral{\prod_{\timecurr' = \timeprev + 1}^{\timecurr} \pdftrans{\param^{\timecurr'}}{\param^{\timecurr'-1}}}{\param^{\timeprev+1} \cdots}{}}{\param^{\timecurr-1}}{} \nonumber
\end{align}
The third term in Equation~\ref{eqn.prior2} corresponds to atoms that have been
instantiated at the current time $\timecurr$. In this case, we know both that the atom survived
and its current value, so $\survive$ and $\trans$ disappear.
Also, the count $\numcount^{\cluster, \leq \timecurr}$
now includes cluster assignments at the current time $\timecurr$ as well.



\section{Derivation of closed-form inference expressions for application of DDPs to world modeling
(Section~\ref{sec.ddp-worldmodel})}
\label{sec.appendix-worldmodel}

In this appendix, we derive closed-form expressions for
the posterior and predictive distributions of the parameter $\param = \parens{\attr, \pose}$,
under the assumptions specified in Section~\ref{sec.ddp-worldmodel}.

The expressions for the fixed attribute are the same as in \cite{Wong2015} \todo{the previous chapter},
since it is static. For convenience, we reproduce the equations here.
Given a set of observations $\braces{\attrobs}$:
\begin{align}
\label{eqn.attr-posterior}
\belpmf(\attr) \triangleq \probcond{\attr}{\braces{\attrobs}} &\propto
\probcond{\braces{\attrobs}}{\attr} \; \prob \parens{\attr} \propto
\brackets{\prod_{\attrobs_i \in \braces{\attrobs}} \attrnoisefn{\attr}{\attrobs_i}} \; \attrprior(\attr) \\
\label{eqn.attr-predictive}
\probcond{\attrobs'}{\braces{\attrobs}} &\propto
\sum_{\attr} \probcond{\attrobs'}{\attr} \; \probcond{\attr}{\braces{\attrobs}} =
\sum_{\attr} \attrnoisefn{\attr}{\attrobs'} \; \belpmf(\attr)
\end{align}

\todo{Better notation than $\numcount^\timecurr$?}
Given a set of observations $\braces{\braces{\worlddef{\poseobs}}_{\defind=1}^{\numcount^\timecurr}}_{\timecurr = \timebirth}^{\timedeath}$
of the dynamic attributes, we can find the posterior distribution on
$\braces{\pose^\timecurr}_{\timecurr = \timebirth}^{\timedeath}$
by performing Kalman filtering and smoothing.
Applying a generic Kalman filter to the world modeling problem gives the following
recursive filtering equations for $\parens{\tilde{\belmean}, \tilde{\belcov}}$,
the hyperparameters in the forward direction (during filtering):
\begin{alignat}{2}
\label{eqn.kalmanfilter}
\hat{\belmean}^\timecurr &= \tilde{\belmean}^{\timecurr-1} \;,\;
&&\hat{\belcov}^\timecurr = \tilde{\belcov}^{\timecurr-1} + \transcov(\attr) \nonumber \\
K^\timecurr &= \begin{cases}
\hat{\belcov}^\timecurr \parens{\hat{\belcov}^\timecurr + \frac{\sensecov}{\numcount^\timecurr}}^{-1} \;,
& \numcount^\timecurr > 0 \\
\mathbf{0} \;, & \numcount^\timecurr = 0
\end{cases} \\
\tilde{\belmean}^\timecurr &= \hat{\belmean}^\timecurr + K^\timecurr \parens{\bar{\poseobs}^\timecurr - \hat{\belmean}^\timecurr} \;,\;
&&\tilde{\belcov}^\timecurr = \parens{I - K^\timecurr} \hat{\belcov}^\timecurr \nonumber
\end{alignat}
\todo{Remind again that we assume most-likely version of dynamics?}
Recall that $\transcov(\attr)$ is the covariance per time step of the random walk on $\pose$,
and $\sensecov$ is the covariance of the measurement noise distribution.
The ``\^{}'' variables are the predicted parameters before incorporating observations,
and the ``\~{}'' variables are the parameters after incorporating observations
(i.e., the Kalman filter output).
Since there may be multiple observations of the pose in a single epoch,
we have used an equivalent formulation involving the sample means $\bar{\poseobs}$,
by exploiting the fact that if each $\worlddef{\poseobs} \sim \normal \parens{\pose^\timecurr, \sensecov}$,
then the sample mean has distribution
$\bar{\poseobs}^\timecurr \sim \normal \parens{\pose^\timecurr, \frac{\sensecov}{\numcount^\timecurr}}$.
There may also be no observations at a given time, in which case the correction step has no effect
($K^\timecurr = 0$).

The Kalman filter is initialized with a noninformative prior:
\begin{align}
\belmean^0 = \mathbf{0} \;,\; \belcov^0 = \infty I 
\end{align}
In practice, this implies that after the initial measurement(s) at time $\timebirth$,
$\tilde{\pose}^\timebirth \sim \normal \parens{\bar{\poseobs}^\timebirth, \frac{\sensecov}{\numcount^\timebirth}}$.
To see this, we can apply Equation~\ref{eqn.kalmanfilter} on $\parens{\belmean^0, \belcov^0}$:
\begin{align}
K^\timebirth &= \parens{\belcov^0 + \transcov(\attr)}
\parens{\belcov^0 + \transcov(\attr) + \frac{\sensecov}{\numcount^\timebirth}}^{-1} \\
\tilde{\belmean}^\timebirth &= \belmean^0 + K^\timebirth \parens{\bar{\poseobs}^\timebirth - \belmean^0}
= K^\timebirth \bar{\poseobs}^\timebirth \\
\tilde{\belcov}^\timebirth &= \parens{I - K^\timebirth} \parens{\belcov^0 + \transcov(\attr)}
\end{align}
To handle the infinite initial covariance, we interpret $\belcov^0$ as $\lim_{n \rightarrow \infty} nI$.
This leads to:
\begin{align}
K^\timebirth &= \lim_{n \rightarrow \infty} \brackets{
nI \parens{nI + \transcov(\attr) + \frac{\sensecov}{\numcount^\timebirth}}^{-1}
+ \transcov(\attr) \parens{nI + \transcov(\attr) + \frac{\sensecov}{\numcount^\timebirth}}^{-1}} \nonumber \\
&= \lim_{n \rightarrow \infty} \brackets{
\parens{I + \frac{\transcov(\attr)}{n} + \frac{1}{n} \frac{\sensecov}{\numcount^\timebirth}}^{-1}
+ \frac{1}{n} \transcov(\attr)
\parens{I + \frac{\transcov(\attr)}{n} + \frac{1}{n} \frac{\sensecov}{\numcount^\timebirth}}^{-1}} \nonumber \\
&= I + 0 \cdot \transcov(\attr) \cdot I = I
\end{align}
Hence $\tilde{\belmean}^\timebirth = K^\timebirth \bar{\poseobs}^\timebirth = \bar{\poseobs}^\timebirth$.
For the covariance:
\begin{align}
\tilde{\belcov}^\timebirth &= \brackets{I - \parens{\belcov^0 + \transcov(\attr)}
\parens{\belcov^0 + \transcov(\attr) + \frac{\sensecov}{\numcount^\timebirth}}^{-1}}
\parens{\belcov^0 + \transcov(\attr)} \nonumber \\
&= \brackets{\parens{\belcov^0 + \transcov(\attr) + \frac{\sensecov}{\numcount^\timebirth}}
\parens{\belcov^0 + \transcov(\attr) + \frac{\sensecov}{\numcount^\timebirth}}^{-1}
- \parens{\belcov^0 + \transcov(\attr)}
\parens{\belcov^0 + \transcov(\attr) + \frac{\sensecov}{\numcount^\timebirth}}^{-1}}
\parens{\belcov^0 + \transcov(\attr)} \nonumber \\
&= \frac{\sensecov}{\numcount^\timebirth}
\parens{\belcov^0 + \transcov(\attr) + \frac{\sensecov}{\numcount^\timebirth}}^{-1}
\parens{\belcov^0 + \transcov(\attr)} \nonumber \\
&= \lim_{n \rightarrow \infty} \brackets{
\frac{\sensecov}{\numcount^\timebirth} \parens{nI + \transcov(\attr) + \frac{\sensecov}{\numcount^\timebirth}}^{-1} nI
+ \frac{\sensecov}{\numcount^\timebirth} \parens{nI + \transcov(\attr) + \frac{\sensecov}{\numcount^\timebirth}}^{-1} \transcov(\attr)}
\nonumber \\
&= \lim_{n \rightarrow \infty} \brackets{
\frac{\sensecov}{\numcount^\timebirth} \parens{I + \frac{\transcov(\attr)}{n} + \frac{1}{n} \frac{\sensecov}{\numcount^\timebirth}}^{-1}
+ \frac{1}{n} \frac{\sensecov}{\numcount^\timebirth} \parens{I + \frac{\transcov(\attr)}{n} + \frac{1}{n} \frac{\sensecov}{\numcount^\timebirth}}^{-1} \transcov(\attr)} \nonumber \\
&= \frac{\sensecov}{\numcount^\timebirth} \cdot I + 0 \cdot \frac{\sensecov}{\numcount^\timebirth} \cdot I \cdot \transcov(\attr)
= \frac{\sensecov}{\numcount^\timebirth}
\end{align}
In summary, choosing $\parens{\belmean^0, \belcov^0} = \parens{\mathbf{0}, \infty I}$
is equivalent to initializing the Kalman filter with
$\parens{\belmean^\timebirth, \belcov^\timebirth} = \parens{\bar{\poseobs}^\timebirth, \frac{\sensecov}{\numcount^\timebirth}}$,
and proceeding for times $\timebirth < t \leq \timedeath$.

After proceeding forward in time, information from later observations should also be propagated
\emph{backward} in time via a smoothing operation. For example, for our application,
the Rauch-Tung-Striebel (RTS) smoother runs the following recursive operations starting at time $\timedeath$:
\begin{alignat}{2}
\belmean^\timedeath &= \tilde{\belmean}^\timedeath \;,\; \belcov^\timedeath = \tilde{\belcov}^\timedeath && \\
C^\timecurr &= \tilde{\belcov}^\timecurr \parens{\hat{\belcov}^{\timecurr+1}}^{-1}
&&= \tilde{\belcov}^\timecurr \parens{\tilde{\belcov}^\timecurr + \transcov(\attr)}^{-1} \\
\belmean^\timecurr &= \tilde{\belmean}^\timecurr + C^\timecurr \parens{\belmean^{\timecurr+1} - \hat{\belmean}^{\timecurr+1}}
&&= \tilde{\belmean}^\timecurr + C^\timecurr \parens{\belmean^{\timecurr+1} - \tilde{\belmean}^\timecurr} \\
\belcov^\timecurr &= \tilde{\belcov}^\timecurr +
C^\timecurr \parens{\belcov^{\timecurr+1} - \hat{\belcov}^{\timecurr+1}} \parens{C^\timecurr}^\intercal
&&= \tilde{\belcov}^\timecurr + C^\timecurr \parens{\belcov^{\timecurr+1}
- \tilde{\belcov}^\timecurr - \transcov(\attr)} \parens{C^\timecurr}^\intercal
\end{alignat}
Recall that ``\^{}'' and ``\~{}'' variables are the predicted and filtered parameters respectively.
Parameters without such modifications are smoothed.

Once the sequence of parameters $\braces{\belmean^\timecurr, \belcov^\timecurr}_{\timecurr = \timebirth}^\timedeath$
is inferred, we can use them to determine the log-likelihood of the observations (for scoring associations)
and the predictive distributions (for determining cluster assignment in Gibbs sampling).
We will repeatedly use the following fact:
\begin{align}
\label{eqn.normal-marginal}
x \sim \normal \parens{\mu, \Sigma} \;,\; \left. y \middle| x \right. \sim \normal \parens{x, \Lambda} \;\Rightarrow\;
y \sim \integral{\probcond{y}{x} \; \prob \parens{x}}{x}{}
= \normal \parens{\mu, \Sigma + \Lambda}
\end{align}
For example, we know that $\pose^\timecurr \sim \normal \parens{\belmean^\timecurr, \belcov^\timecurr}$
(hyperparameters obtained from Kalman smoothing), and from our modeling assumptions,
$\left. \poseobs^\timecurr \middle| \pose^\timecurr \right. \sim \normal \parens{\pose^\timecurr, \sensecov}$.
Hence the marginal distribution over the pose observation
(marginalized over all possible latent poses $\pose^\timecurr$) is
$\poseobs^\timecurr \sim \normal \parens{\belmean^\timecurr, \belcov^\timecurr + \sensecov}$.
From this we can immediately find the marginal likelihood of the observed data:
\begin{align}
\label{eqn.marginal-likelihood}
\prob \parens{\braces{\braces{\worlddef{\poseobs}}_{\defind=1}^{\numcount^\timecurr}}_{\timecurr = \timebirth}^{\timedeath}}
= \prod_{\timecurr=\timebirth}^\timedeath \prod_{\defind=1}^{\numcount^\timecurr} \prob \parens{\poseobs_\defind^\timecurr}
= \prod_{\timecurr=\timebirth}^\timedeath \prod_{\defind=1}^{\numcount^\timecurr}
\pdfnormal{\poseobs_\defind^\timecurr}{\belmean^\timecurr, \belcov^\timecurr + \sensecov}
\end{align}
\review{This likelihood expression is used to score potential association hypotheses.}

We can now derive the conditional probability expressions in the Gibbs sampler,
shown in Equation~\ref{eqn.forward-collapsed}.
In collapsed Gibbs sampling, each observation's predictive likelihood
$\probcond{\worlddef{\obs}}{\excludedef{\Obs}^\cluster}$
involves an integral over the latent parameters
$\beldef{\param} = \parens{\attr^\cluster, \beldef{\pose}}$ of the cluster.
In forward sampling, assigning observation $\worlddef{\obs}$ to cluster $\cluster$ has three cases:
\begin{enumerate}
\item If cluster $\cluster$ exists and is instantiated
(i.e., has other observations at time $\timecurr$ assigned to it),
the posterior distribution of the pose $\beldef{\pose}$ is
$\normal \parens{\beldef{\belmean}, \beldef{\belcov}}$,
and the posterior distribution of the fixed attribute is $\belpmf(\beldeffixed{\attr})$.
Thus the predictive distribution is:
\begin{align}
\label{eqn.gibbs-case1}
\probcond{\worlddef{\obs}}{\excludedef{\Obs}^\cluster}
&= \integral{\probcond{\worlddef{\obs}}{\param} \; \probcond{\param}{\excludedef{\Obs}^\cluster}}{\param}{}
\nonumber \\
&= \brackets{\sum_{\beldeffixed{\attr}} \probcond{\worlddef{\attrobs}}{\beldeffixed{\attr}} \;
\belpmf(\beldeffixed{\attr})} \;
\integral{\probcond{\worlddef{\poseobs}}{\beldef{\pose}} \;
\pdfnormal{\beldef{\pose}}{\beldef{\belmean}, \beldef{\belcov}}}{\beldef{\pose}}{} \nonumber \\
&= \brackets{\sum_{\beldeffixed{\attr}} \attrnoisefn{\beldeffixed{\attr}}{\worlddef{\attrobs}} \;
\belpmf(\beldeffixed{\attr})} \;
\pdfnormal{\worlddef{\poseobs}}{\beldef{\belmean}, \beldef{\belcov} + \sensecov}
\end{align}
In the final line, we used Equations~\ref{eqn.attr-predictive} and \ref{eqn.normal-marginal}
to simplify the predictive distributions.
\item If cluster $\cluster$ exists, but it has not yet been instantiated,
this implies that, in the forward case, that the time of the observation $\timecurr$
is beyond the final observed time $\timeprev = \timedeath$ associated with the cluster.
Then instead of integrating over the posterior distribution of $\beldef{\pose}$,
which does not exist yet, we need to integrate over its \emph{predictive} distribution.
This can be found by propagating the prediction step in the Kalman filter,
starting from the final time step's distribution
$\beldefprev{\pose} \sim \normal \parens{\beldefprev{\belmean}, \beldefprev{\belcov}}$:
\begin{align}
\beldef{\belmean} = \beldefprev{\belmean} \;,\;
\beldef{\belcov} = \beldefprev{\belcov} + (\timecurr - \timeprev) \transcov(\beldeffixed{\attr})
\end{align}
This is precisely the ``generalized'' transition distribution $\transgen$ for the pose in the DDPMM.
Hence $\beldef{\pose} \sim \normal \parens{\beldefprev{\belmean},
\beldefprev{\belcov} + (\timecurr - \timeprev) \transcov(\beldeffixed{\attr})}$,
and by a derivation similar to Equation~\ref{eqn.gibbs-case1}:
\begin{align}
\label{eqn.gibbs-case2}
\probcond{\worlddef{\obs}}{\excludedef{\Obs}^\cluster}
&= \brackets{\sum_{\beldeffixed{\attr}} \attrnoisefn{\beldeffixed{\attr}}{\worlddef{\attrobs}} \;
\belpmf(\beldeffixed{\attr})} \;
\pdfnormal{\worlddef{\poseobs}}{\beldefprev{\belmean},
\beldefprev{\belcov} + (\timecurr - \timeprev) \transcov(\beldeffixed{\attr}) + \sensecov}
\end{align}
\item If cluster $\cluster$ does not exist (and $\excludedef{\Obs}^\cluster = \emptyset$),
then we should use the base distribution
$\pdfbase{\param} \triangleq \attrprior(\attr) \; \pdfnormal{\pose}{\belmean^0, \belcov^0}$
instead of the posterior distribution. Then:
\begin{align}
\probcond{\worlddef{\obs}}{\excludedef{\Obs}^\cluster} &= \prob \parens{\worlddef{\obs}}
= \integral{\probcond{\worlddef{\obs}}{\param} \; \pdfbase{\param}}{\param}{}
\nonumber \\
&= \brackets{\sum_{\beldeffixed{\attr}} \probcond{\worlddef{\attrobs}}{\beldeffixed{\attr}} \;
\attrprior(\beldeffixed{\attr})} \;
\integral{\probcond{\worlddef{\poseobs}}{\beldef{\pose}} \;
\pdfnormal{\beldef{\pose}}{\belmean^0, \belcov^0}}{\beldef{\pose}}{} \nonumber \\
&= \brackets{\sum_{\beldeffixed{\attr}} \attrnoisefn{\beldeffixed{\attr}}{\worlddef{\attrobs}} \;
\attrprior(\beldeffixed{\attr})} \;
\pdfnormal{\worlddef{\poseobs}}{\mathbf{0}, \infty I}
\end{align}
However, this requires the evaluation of an \review{improper} normal distribution in the final term.
Since the choice of this prior was motivated by an attempt to give all initial poses equal probability,
the same effect can be achieved by using a uniform distribution over the total explored world volume.
Thus, in practice during Gibbs sampling we evaluate the following:
\begin{align}
\label{eqn.gibbs-case5}
\probcond{\worlddef{\obs}}{\excludedef{\Obs}^\cluster} &=
\brackets{\sum_{\beldeffixed{\attr}} \attrnoisefn{\beldeffixed{\attr}}{\worlddef{\attrobs}} \;
\attrprior(\beldeffixed{\attr})} \;
\unif(\volume(\mathrm{world}))
\end{align}
\review{Note that this is similar to the expression for the observation likelihood of false positives.}
If the observation is actually assigned to a new cluster,
then we revert to the noninformative normal prior and perform Kalman smoothing,
which is now no longer problematic since it does not require evaluation of \review{improper} densities.
\end{enumerate}

\newpage

\bibliographystyle{plainnat}
\bibliography{references}

\end{document}